\documentclass[manuscript,screen]{acmart}

\AtBeginDocument{%
  \providecommand\BibTeX{{%
    \normalfont B\kern-0.5em{\scshape i\kern-0.25em b}\kern-0.8em\TeX}}}
\usepackage{subfigure}
\usepackage{wrapfig}
\usepackage[ruled, linesnumbered]{algorithm2e}
\usepackage{soul}
\def\model{VTFS}

\setcopyright{acmlicensed}
\copyrightyear{2024}
\acmYear{2024}
\acmDOI{XXXXXXX.XXXXXXX}

\acmISBN{978-1-4503-XXXX-X/18/06}

\begin{document}

\title{Feature Selection as Deep Sequential Generative Learning}


\author{Wangyang Ying}
\affiliation{
  \institution{Arizona State University, School of Computing and Augmented Intelligence, Tempe}
  \country{USA}}
\email{yingwangyang@gmail.com}

\author{Dongjie Wang}
\affiliation{%
  \institution{Department of Computer Science, University of Kansas, Lawrence}
  \country{USA}
}\email{wangdongjie@ku.edu}

\author{Haifeng Chen}
\affiliation{%
 \institution{NEC Laboratories America Inc, Princeton}
 \country{USA}}
\email{ haifeng@nec-labs.com}

\author{Yanjie Fu}
\affiliation{%
  \institution{Arizona State University, School of Computing and Augmented Intelligence, Tempe}
  \country{USA}}
\email{yanjie.fu@asu.edu}

\renewcommand{\shortauthors}{Ying et al.}

\begin{abstract}
Feature selection aims to identify the most pattern-discriminative feature subset. 
In prior literature, 
filter (e.g., backward elimination) and embedded (e.g., Lasso) methods have hyperparameters (e.g., top-K, score thresholding) and tie to specific models, thus, hard to generalize; wrapper methods search a feature subset in a huge discrete space and is computationally costly.  
To transform the way of feature selection, 
we regard a selected feature subset as a selection decision token sequence and reformulate feature selection as a deep sequential generative learning task that distills feature knowledge and generates decision sequences. 
Our method includes three steps: 
(1) We develop a deep variational transformer model over a joint of sequential reconstruction, variational, and performance evaluator losses. Our model can distill feature selection knowledge and learn a continuous embedding space to map feature selection decision sequences into embedding vectors associated with utility scores. 
(2) We leverage the trained feature subset utility evaluator as a gradient provider to guide the identification of the optimal feature subset embedding;
(3) We decode the optimal feature subset embedding to autoregressively generate the best feature selection decision sequence with autostop. 
Extensive experimental results show this generative perspective is effective and generic, without large discrete search space and expert-specific hyperparameters. 
The code is available at \url{http://tinyurl.com/FSDSGL}

\end{abstract}





\maketitle

\section{Introduction}

Feature selection aims to identify the best feature subset from an original feature set. 
Effective feature selection methods reduce dataset dimensionality, shorten training time, prevent overfitting,  enhance generalization, and, moreover, improve the performance of downstream machine learning tasks.
The applicability of this technique can be applied to multiple domains, including biomarker discovery, traffic forecasting, financial analysis, urban computing, etc.

Prior literature can be categorized as: 
1) Filter methods~\cite{kbest,forman2003extensive,hall1999feature,yu2003feature}  rank features based on a score (e.g., relevance between feature and label) and select top-$k$ features as the optimal feature subset (e.g., univariate feature selection). 
2) Embedded methods~\cite{lasso,sugumaran2007feature} jointly optimize feature selection and downstream prediction tasks. For instance, LASSO shrinks feature coefficients by optimizing regression and regularization loss. 
3) Wrapper methods~\cite{yang1998feature,kim2000feature,narendra1977branch,kohavi1997wrappers} formulate feature selection as a searching problem in a large discrete feature combination space via evolutionary algorithms or genetic algorithm that collaborate with a downstream machine learning model. 

However, existing studies are not sufficient. Filter methods typically overlook relationships between features, are sensitive to data distribution, and are non-learnable, hence they often perform poorly.
Embedded methods rely on strong structured assumptions (e.g., sparse coefficients of L norm) and downstream models (e.g., regression), making them inflexible. Wrapper methods suffer from exponentially growing discrete search space (e.g., around $2^N$ if the feature number is N). 
Can we develop a more effective learning framework without searching a large discrete space? 

\begin{figure}[h]
    \centering
    \subfigure[Iterative selection view]{
    \begin{minipage}[ht]{0.45\linewidth}
    \centering
    \includegraphics[width=0.5\linewidth]{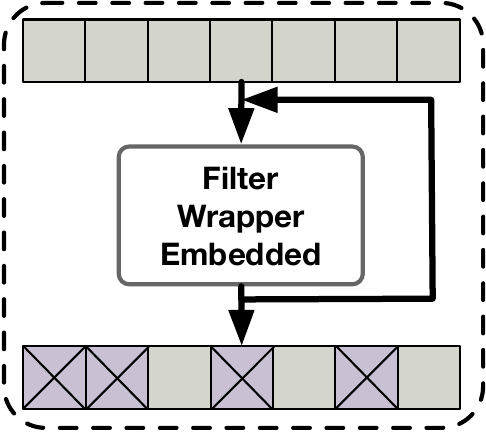}
    \label{method_contrast_1}
    \end{minipage}
    }
    \hspace{-3cm}
    \centering
    \subfigure[Generative view]{
    \begin{minipage}[ht]{0.5\linewidth}
    \centering
    \includegraphics[width=0.45\linewidth]{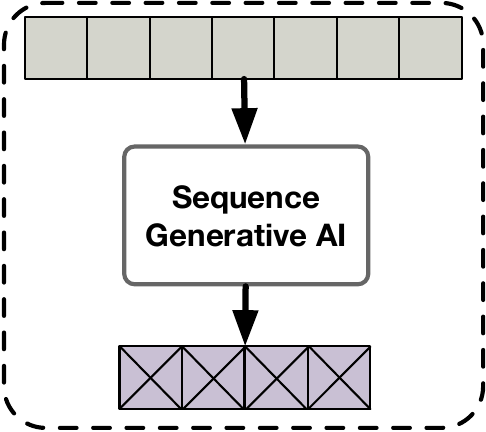}
    \label{method_contrast_2}
    \end{minipage}
    }
    \caption{Our perspective can be viewed as a sequence generation (b) rather than as an iterative discrete selection (a).} \label{method_contrast}
\end{figure}

\textbf{Our Perspective: Feature Selection as Sequential Generative AI.}
The emerging Artificial Generative Intelligence (AGI) and ChatGPT show it is possible to learn complex and mechanism-unknown knowledge from historical experiences and make smart decisions in an autoregressive generative fashion. 
Following a similar spirit, we believe that knowledge related to feature subsets can also be distilled and embedded into a continuous space, where computation and optimization are enabled and, thereafter, generate a feature selection decision sequence. 
This generative perspective regards feature selection,  e.g., $f_1f_2,...,f_7 \rightarrow f_1f_2f_4f_6$, as a sequential generative learning task to generate an autoregressive feature selection decision sequence (\textbf{Figure~\ref{method_contrast_2}}). This transforms the traditional way we select features via an iterative subset selection process (\textbf{Figure~\ref{method_contrast_1}}).
Under this generative perspective,  a feature subset is represented as a feature token sequence and subsequently embedded in a differentiable continuous space.  
In this continuous embedding space,  an embedding vector corresponds to a feature subset, and we can: a) build an evaluator function to assess feature subset utility; b)  search the optimal feature subset embedding;  c) decode an embedding vector to generate a feature selection decision sequence. 
This generative learning perspective provides great potential to distill feature knowledge from experiences and generalize well over various domain datasets.

Inspired by these findings, we propose a \textbf{deep \underline{v}ariational sequen\underline{t}ial generative \underline{f}eature \underline{s}election learning} (\textbf{VTFS}) framework that includes three steps:
1) \ul{\emph{Embedding.}} 
We develop a variational transformer model with joint optimization of sequence reconstruction loss,  feature subset accuracy evaluator loss, and variational distribution alignment (i.e., Kullback–Leibler) loss, in order to learn a feature subset embedding space. This strategy can strengthen the ability of model denoising and reduce noise feature selection. 
2) \ul{\emph{Optimization.}} 
After the convergence of the embedding space, we leverage the evaluator to generate gradient and direction information, enabling us to effectively steer gradient ascent-based search and identify the embedding for the optimal feature subset. 
3) \ul{\emph{Generation.}}  
We decode the optimal embedding and autoregressively generate the optimal feature token sequence. Finally, we apply the optimal feature token sequence to the original feature set to get the best feature subset. 
In addition, to prepare historical feature selection experiences and corresponding model performance as training data,  we leverage the automation and exploration properties of reinforcement intelligence to develop a training data collector. The collector can explore and collect feature subset-predictive accuracy pairs as training data. This strategy can avoid intensive manual labor, and improve training data quality and diversity. 

Our main contributions can be summarized as follows:
\begin{enumerate}
    \item \emph{Generative perspective:} We propose a formulation: feature selection as deep sequential generative AI to convert the discrete selection process into continuous optimization. 

    \item \emph{EOG (Embedding-Optimization-Generation) framework:} We develop the EOG framework: embedding feature subsets to vectors, gradient-steered optimal embedding identification, and feature token sequence generation. Extensive experiments show that this generative framework improves the effectiveness and generalization of feature selection in various data domains.  

    \item \emph{Computing Techniques:} We design interesting techniques to address computing issues: a) reinforcement as an automated feature selection training data collector, b) variational transformer with multi-losses as optimization supervision, and c) performance evaluator function as gradient generator.

    \item \emph{Extensive Experiments:} We conduct extensive experiments and case studies across 16 real-world datasets to demonstrate the effectiveness, robustness, and scalability of our framework.
\end{enumerate}

\section{Preliminaries and Problem Statement}
\noindent\textbf{Feature Token Sequence.}
We formulate a feature subset as a feature token sequence so that we can encode it into an embedding space with a deep sequential model. Specifically, we treat each feature as a token and construct a mapping table between features and tokens. For example,  given a feature subset $[f_1, f_2, f_4, f_7]$, we convert it to a feature token sequence denoted as $[SOS, t_1, t_2, t_4, t_7, EOS]$. 

\noindent\textbf{Sequential Training Data.}
To construct a differential embedding space for feature selection, we need to collect $N$ different feature subset-accuracy pairs from the original feature set as training data. Then we convert all feature subsets to feature token sequences.
These data is denoted by $R = (\mathbf{t}_i, v_i)^N_{i=1}$, where $\mathbf{t}_i = [t_1, t_2, ..., t_q ]$ is the feature token sequence of the $i$-th feature subset, and $v_i$ is corresponding downstream predictive accuracy.

\noindent\textbf{Problem Statement.}
Formally, given a tabular data set $D = (X, y)$, where $X$ is an original feature set and y is the corresponding target label. We collect the sequential training data $R$ by conducting automatically traditional feature selection algorithms on $D$ and evaluating the performance of feature subsets with a downstream machine learning model.
Our goal is to 1) embed the knowledge of $R$ into a differentiable continuous space and 2) generate the optimal feature subset. Regarding goal 1, we learn an encoder $\phi$, an evaluator $\vartheta$, and a decoder $\psi$ via joint optimization to get the feature subset embedding space $\mathcal{E}$. Regarding goal 2, we identify the best embedding based on a gradient search method and generate the optimal feature token sequence $\mathbf{t}^*$:
\begin{equation}
    \mathbf{t}^* = \psi(E^*) = \arg\max_{E \in \mathcal{E}}\mathcal{M}(X[\psi(E)], y), 
\end{equation}
where $\psi$ is a decoder to generate a feature token sequence from any embedding of $\mathcal{E}$; $E^*$ is the optimal feature subset embedding; $\mathcal{M}$ is a downstream ML task. $X[]$ means we use the mapping table to convert a feature token sequence to a feature subset. Finally, we apply $\mathbf{f}^*$ to $X$ to select the optimal feature subset $X[\mathbf{t}^*]$. 
\section{Methodology}

\subsection{Framework Overview}

\begin{figure*}[]
        \centering
	\includegraphics[width=1.0\linewidth]{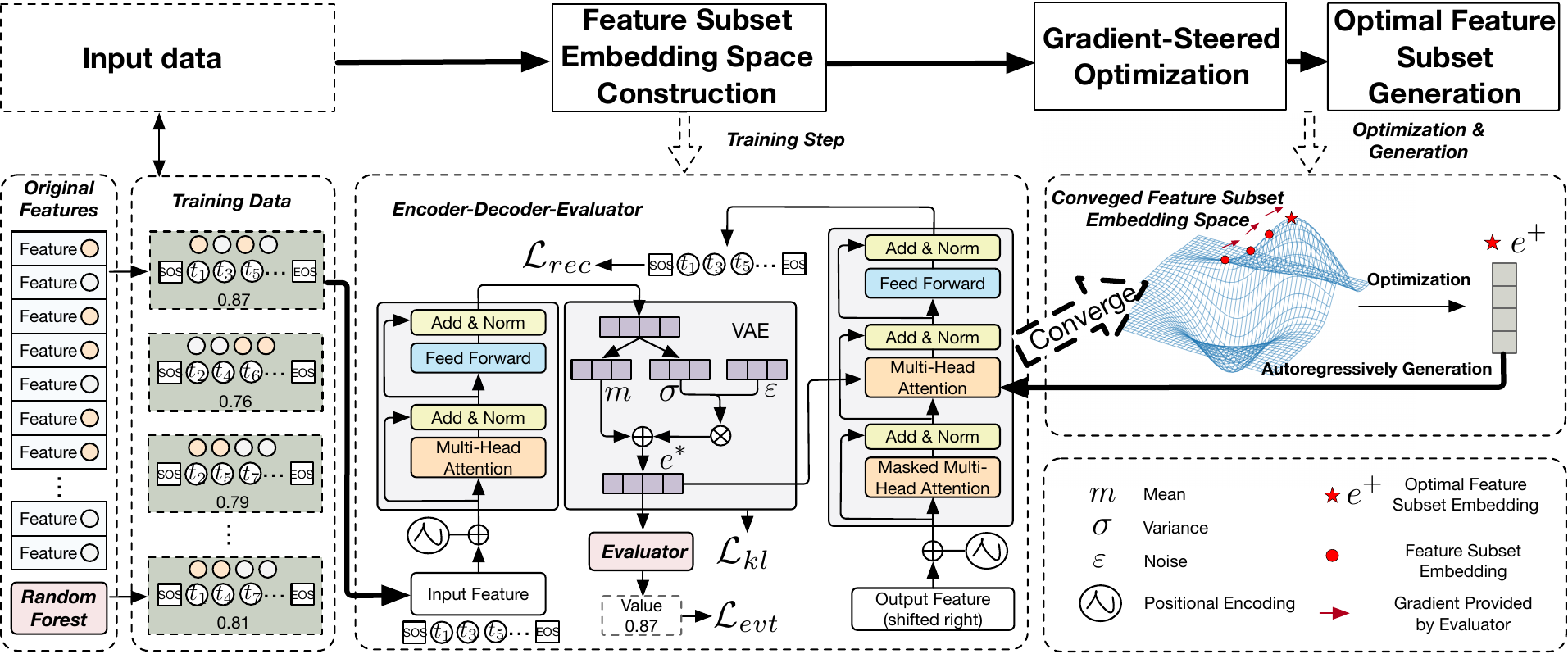}
	\caption{An overview of VTFS. First, we employ the variational transformer-based sequential model to construct feature subset embedding space. Second, we search for better embeddings by moving local optimal embeddings along the gradient direction maximizing the downstream predictive accuracy. Third, we generate the feature token sequences in an autoregressive manner based on these better embeddings and keep the best one with the highest downstream ML performance.}
	\label{framework}
    % \vspace{-0.3cm}
\end{figure*}

Figure~\ref{framework} shows that our framework (VTFS), which includes three steps:  
1) feature subset embedding space construction, 
2) gradient-steered optimization, 
and 3) optimal feature subset generation. 
Specifically, Step 1 is to embed the knowledge of feature selection into a continuous embedding space.
To accomplish this, we develop an encoder-decoder-evaluator architecture, in which the encoder encodes each feature token sequence into an embedding vector; the evaluator estimates the downstream prediction task accuracy based on the corresponding embedding; the decoder reconstructs the associated feature token sequence using the embedding. 
To construct a distinguishable and smooth embedding space, we employ a variational transformer as the backbone of the sequential model.
We jointly optimize the sequence reconstruction loss and the performance estimation loss to learn such an embedding space.
Then, we employ the gradient-steered search to find the better embeddings in Step 2.
We select the top K feature token sequence from the collected data based on predictive accuracy.
They are converted into embeddings using the well-trained encoder.
After that, based on the gradient of the well-trained evaluator, we move these embeddings along the direction maximizing the downstream task performance to get better ones.
Finally, in Step 3, we feed the better embeddings into the well-trained decoder to generate the feature token sequences and then convert them to the feature subsets.
The feature subset with the highest downstream ML performance is regarded as the optimal result.

\subsection{Feature Subset Embedding Space Construction via Variational Transformer}
The success of ChatGPT illustrates that intricate human knowledge can be effectively embedded within a large embedding space via sequential modeling.
This inspiration encourages that feature selection, as a form of human knowledge, can likewise be integrated into a continuous embedding space.
However, different from ChatGPT,  we expect this embedding space should not only preserve the knowledge of feature subsets but also maintain the quality of these subsets. 
This is crucial for the effective identification of the optimal feature selection result.
To achieve this, we develop an encoder-decoder-evaluator learning paradigm.

\textbf{Feature subsets as sequences with shuffling-based augmentations.} 
The sequential training data is used to construct the continuous embedding space.
We find that the order of the feature token sequence doesn't influence the predictive accuracy. 
Thus, we propose a shuffling-based strategy to quickly collect more legal data samples.
For instance, give one sample $[t_1, t_2, t_3]\rightarrow0.867$. 
We can shuffle the order of the sequence to generate more semantically equivalent data samples: $[t_2,t_1,t_3]\rightarrow0.867$, $[t_3,t_2,t_1]\rightarrow0.867$. 
The shuffling augmentation strategy enhances both the volume and diversity of data, enabling the construction of an empirical training set that more accurately represents the true population. This strategy is significant in developing a more effective continuous embedding space.

\textbf{Variational transformer-based feature subset embedding model.} 
We develop an encoder-decoder-evaluator framework to embed complex feature learning knowledge into a continuous embedding space.
Such a space should preserve the influence of different feature subsets, while also maintaining a smooth structure to facilitate the identification of superior embeddings.
To accomplish this, we adopt the variational transformer~\cite{vaswani2017attention,vae} as the backbone of the sequential model to implement this framework.

\underline{\textit{The Encoder}} aims to embed a feature token sequence into an embedding vector.  
Formally, consider a training dataset $R = {(\mathbf{t}_i, v_i)}_{i=1}^N $, where $\mathbf{t}_i = [t_1, t_2, ..., t_q]$ is a feature token sequence of the $i$-th feature subset, $v_i$ is the corresponding predictive accuracy of, $q$ is the number tokens of the $i$-th feature token sequence, and $N$ is the number of training samples.
To simplify the notation, we use the notation $(\mathbf{t}, v)$ to represent any training sample.
We first employ a transformer encoder $\phi$  to learn the embedding of the feature token sequence, denoted by  $\mathbf{e} = \phi(\mathbf{t})$. 
We assume that the learned embeddings $\mathbf{e}$ follow the format of normal distribution.
Then, two fully connected layers are implemented to estimate the mean $\mathbf{m}$ and variance $\mathbf{\sigma}$ of this distribution.
After that, we can sample an embedding vector $\mathbf{e^*}$ from the distribution via the reparameterization technique.
This process is denoted by 
\begin{equation}
   \mathbf{e^*} = \mathbf{m} + \mathbf{\varepsilon} * exp(\mathbf{\sigma}), 
\end{equation}
where $\varepsilon$ refers to the noised vector sampled from a standard normal distribution.
The sampled vector $\mathbf{e^*}$ is regarded as the input of the following decoder and evaluator.

\underline{\textit{The Decoder}} aims to reconstruct a feature token sequence using the embedding $\mathbf{e^*}$.  
We utilize a transformer decoder to parse the information of $\mathbf{e^*}$ and add a softmax layer behind it to estimate the probability of the next feature token based on the previous ones.
Formally, the current token that needs to be decoded is $t_j$, and the previously completed feature token sequence is $t_1...t_{j-1}$. 
The probability of the $j$-th token should be:
\begin{equation}
    P_\psi(t_j | \mathbf{e^*}, [t_1,t_2,...,t_{j-1}]) = \frac{exp(z_j)}{\sum_q exp(z)},
\end{equation}
where $z_j$ represents the $j$-th output of the softmax layer, $\psi$ refers to the decoder.
The joint estimated likelihood of the entire feature token sequence should be: 
\begin{equation}
    P_\psi(\mathbf{t} | \mathbf{e^*}) = \prod_{j=1}^q P_\psi(t_j | \mathbf{e^*},[t_1,t_2,...,t_{j-1}])
\end{equation}

\underline{\textit{The Evaluator}} aims to evaluate the predictive accuracy based on the embedding $\mathbf{e^*}$.
More specifically, we implement a fully connected neural layer as the evaluator to predict the corresponding accuracy in the sequential training data.
This calculation process can be denoted by 
\begin{equation}
    \ddot{v} = \vartheta(\mathbf{e^*}), 
\end{equation}
where $\vartheta$ refers to the evaluator and $\ddot{v}$ is the predicted accuracy via $\vartheta$.

\underline{\textit{The Joint Optimization.}} 
We jointly train the encoder, decoder, and evaluator to learn the continuous embedding space. 
There are three objectives: 
a) Minimizing the reconstruction loss between the reconstructed feature token sequence and the real one, denoted by
\begin{equation}
\begin{aligned}
    \mathcal{L}_{rec} &= -logP_\psi(\mathbf{t} | \mathbf{e^*}) \\
                      &= -\sum_{j=1}^qlogP_\psi(t_j | \mathbf{e^*}, [t_1,t_2,...,t_{j-1}]),
\end{aligned}
\end{equation}
b) Minimizing the estimation loss between the predicted accuracy and the real one, denoted by:
\begin{equation}
    \mathcal{L}_{evt} = MSE(v, \ddot{v}),
\end{equation}
c) Minimizing the Kullback–Leibler (KL) divergence between the learned distribution of the feature subset and the standard normal distribution, denoted by:
\begin{equation}
    \mathcal{L}_{kl} = exp(\sigma) - (1 + \sigma) + (m)^2.
\end{equation}
The first two objectives ensure that each point within the embedding space is associated with a specific feature subset and its corresponding predictive accuracy. The last objective smoothens the embedding space, thereby enhancing the efficacy of the following gradient-steered search step. We trade off these three losses and jointly optimize them by:
\begin{equation}
    \mathcal{L} = \alpha\mathcal{L}_{evt}+\beta\mathcal{L}_{rec} + \gamma\mathcal{L}_{kl},
\end{equation}
where $\alpha$, $\beta$ and $\gamma$ are hyper-paramethers.

\subsection{Gradient-steered Optimization}
After obtaining the feature subset embedding space,  we employ a gradient-ascent search 
method to find better feature subset embedding. 
More specifically, we initiate the process by selecting the top $k$ feature token sequences from the collected data based on the corresponding predictive accuracies.
Subsequently, we leverage the encoder that has been well-trained in the last step to convert these feature token sequences into local optimal embeddings
After that, we adopt a gradient-ascent algorithm to move these embeddings along the direction maximizing the downstream predictive accuracy.
The gradient utilized in this process is derived from the well-trained evaluator $\vartheta$.
Taking the embedding $\mathbf{e^*}$ as an illustrative example, the moving calculation process is as follows:
\begin{equation}
    \mathbf{e^+} = \mathbf{e^*} +\eta\frac{\partial\vartheta}{\partial{\mathbf{e^*}}},
\end{equation}
where $\eta$ is the moving steps and $\mathbf{e^+}$ is the better embedding.

\subsection{Optimal Feature Subset Generation}
Once we identify the better embeddings, we will generate the better feature token sequences based on them in an autoregressive manner.
Formally, we take the embedding $\mathbf{e^+}$ as an example to illustrate the generation process. In the $j$-iteration, we assume that the previously generated feature token sequence is $t_1...t_{j-1}$ and the waiting to generate token is $t_j$.
The estimation probability for generating $t_j$ is to maximize the following likelihood based on the well-trained decoder $\psi$:
\begin{equation}
    t_j = \arg\max(P_\psi(t_j | \mathbf{e^+}, [t_1,...,t_{j-1}]).
\end{equation}
We will iteratively generate the possible feature tokens until finding the end token (i.e., $<$EOS$>$).
For instance, if the generated token sequence is ``$[t_2,t_6,t_5,\text{$<$EOS$>$}, t_8]$, '', we will cut from the \text{$<$EOS$>$} token and keep 
$[t_2, t_5, t_6]$ as the final generation result.
Finally, we select the corresponding features according to these feature tokens and output the feature subset with the highest predictive accuracy as the optimal feature subset.
Algorithm~\ref{algo1} shows the pseudo-code of the entire optimization procedure:
\begin{algorithm}
    \SetAlgoNoLine
    \LinesNumbered
    \SetKwInOut{Input}{Input}
    \SetKwInOut{Output}{Output}
    \Input{The original dataset $D = (X, y)$}
    \Output{The Optimal Feature Subset $X[\mathbf{t}^{*}]$ }

    Collecting training data set $R=(\mathbf{t}_i, v_i)_{i=1}^N$. \\
    Initialize the encoder $\phi$, decoder $\psi$ and evaluator $\theta$. \\
    \textbf{Feature Subset Embedding Space Construction:} \\
    \For{$in ~epoch$}
    {
        \For{$in ~number ~of ~batches$}
        {
            Encode: $\mathbf{e} = \phi(\mathbf{t})$. \\
            Estimate: $\mathbf{m}, \mathbf{\sigma}$. \\
            Reparameterization: $\mathbf{e^*} = \mathbf{m} + \varepsilon*exp(\mathbf{\sigma})$. \\
            Decode loss: $\mathcal{L}_{rec} = -logP_{\psi}(\mathbf{t}|\mathbf{e^*})$. \\
            Evaluate loss: $\mathcal{L}_{evt} = MSE(v, \theta(\mathbf{e^*}))$. \\ 
            KL loss: $\mathcal{L}_{kl} = exp(\mathbf{\sigma}) - (1+\mathbf{\sigma}) + (\mathbf{m})^2$. \\
            Backward: $\mathcal{L} = \alpha\mathcal{L}_{evt}+\beta\mathcal{L}_{rec} + \gamma\mathcal{L}_{kl}$
        }
    }
    \textbf{Gradient-steered Optimization:} \\
    Select top-$k$ feature token sequences $(\mathbf{t})^k$ from $R$. \\
    Encode and Reparameterization: $(\mathbf{e^*})^k= reparameterization(\phi((\mathbf{t})^k))$. \\
    Update $(\mathbf{e^*})^k$ with $\eta$ steps: $(\mathbf{e^+})^k = (\mathbf{e^*})^k + \eta*\frac{\partial\vartheta}{\partial{(\mathbf{e^*})^k}}$. \\
    \textbf{Optimal Feature Subset Generation:} \\
    Generation: $(\mathbf{t^+})^k= \psi((\mathbf{e^*})^k)$. \\
    Optimal feature subset: X[$\mathbf{t}^*] = \arg\max\mathcal{M}(X[(\mathbf{t^+})^k], y)$.\\
    \caption{Entire Optimization Procedure}
    \label{algo1}
\end{algorithm}

\begin{figure}
    \centering
    \includegraphics[width=0.5\textwidth]{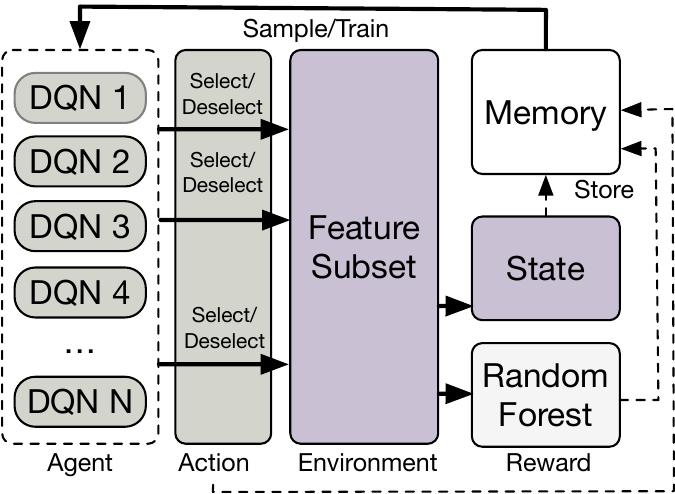}
    \caption{Reinforcement data collector.}
    \label{fig:rl_based_model}
\end{figure}
\subsection{Improvements: Reinforced Data Collector for Sequential Training Data}
To effectively embed feature learning knowledge into an embedding space, we need to explore various feature subsets and collect corresponding predictive accuracy as training data. However, collecting such data requires intensive labor and is time-consuming. Our perspective is to leverage reinforcement intelligence to build a reinforcement data collector to collect diverse, high-quality, and automated feature subset-predictive accuracy pairs as feature learning knowledge training data.
Inspired by~\cite{marlfs}, we believe that the process of feature selection can be modeled by a multi-agent system.
\textbf{Figure \ref{fig:rl_based_model}} shows this system includes two components: 1) reinforcement feature selector; and 2) random forest. 
In particular, to build a reinforcement feature selector, we create an agent for each feature. An agent can take an action to select or deselect the corresponding feature.
We regard the selected feature subset as a reinforcement learning environment. 
So, an action to select or deselect a feature will change the environment. 
The environment will provide two observational feedback: 1) the new environment state after selecting or deselecting a feature; and 2) the predictive accuracy of the downstream random forest model as a reward. 
We categorize agents into participating agents that participate in decision-making to change the feature subset, and non-participating agents that don't change the feature subset. 
In reward assignment, the reward is split equally and then assigned to each participating agent. Non-participating agents receive no reward. 
We use such a personalized reward assignment strategy to incentivize agents to update their selection policy via the value-based learning algorithm of DQN~\cite{DQN}.
The agents have naive policies in the beginning and explore diverse feature subsets with randomness to collect various feature subsets and corresponding random forest accuracy. 
As the agent policies grow, we can collect more high-quality feature subsets with higher accuracy. In this way, we can collect lots of training data samples during the iterative exploration process.
The implementation details of the data collector are included in the code released in the abstract.
\section{Experiments}
\subsection{Experimental Setup}
\setlength{\tabcolsep}{6mm}{
\begin{table}
\centering
\small
\caption{Dataset key statistics. We reported F1-score for classification (C) and 1-RAE for regression (R) respectively.}
\begin{tabular}{cccc}
\toprule \toprule
Dataset          & Task & \#Samples & \#Features \\ \hline
SpectF           & C    & 267       & 44         \\
SVMGuid3         & C    & 1243      & 21         \\
German   Credit  & C    & 1001      & 24         \\
UCI       Credit & C    & 30000     & 25         \\
SpamBase         & C    & 4601      & 57         \\
Ap\_omentum      & C    & 275       & 10936      \\
Ionosphere       & C    & 351       & 34         \\ 
Activity         & C   & 10299     & 561        \\
Mice-Protein     & C   & 1080      & 77         \\\hline
Openml-586       & R    & 1000      & 25         \\
Openml-589       & R    & 1000      & 25         \\
Openml-607       & R    & 1000      & 50         \\
Openml-616       & R    & 500       & 50         \\
Openml-618       & R    & 1000      & 50         \\
Openml-620       & R    & 1000      & 25         \\
Openml-637       & R    & 500       & 50         \\ \bottomrule \bottomrule
\end{tabular}
\label{exp:data_statistics}
\end{table}}
\noindent\textbf{Data Description.} 
We perform experiments using a diverse set of 16 datasets sourced from various domains, including those from UCIrvine and OpenML. These datasets are classified based on their task types into two categories: 1) classification (C) and 2) regression (R). The statistical details of these datasets are presented in Table~\ref{exp:data_statistics}.

\noindent\textbf{Evaluation Design.}
For each of the 16 domain datasets, we randomly constructed two independent data subsets: A and B.
\textbf{Data subset A} was seen by our method. We used this data subset to collect feature subset-accuracy training data pairs (e.g. $f_1f_4f_6 \rightarrow 0.817$) and construct feature subset embedding space.
\textbf{ Data subset B} was never seen by our method. After determining the optimal feature token sequence, such as ${f_2f_5f_6}_A$, using Data subset A, we directly applied this feature token sequence to Data subset B, yielding the feature subset $\{f_2, f_5, f_6\}_B$. This feature subset was used to evaluate the effectiveness of our method.
We use Random Forest as the predictive model for all datasets.
F1-score and 1 - Relative Absolute Error (1- RAE) are regarded as the evaluation metrics for classification and regression tasks respectively.
For the two metrics, the higher the value is, the better the quality of the feature subset is.

\noindent\textbf{Baseline Algorithms.} We compare our method (\textbf{\model}) with 12 widely used feature selection algorithms:
(A). Filter methods:
1) \textbf{K-BEST}~\cite{kbest} selects the top-$k$ features with the highest importance scores; 
2) \textbf{mRMR}~\cite{mrmr} selects a feature subset by maximizing relevance with labels and minimizing feature-feature redundancy; 
3) \textbf{DNP}~\cite{DNP} employs a greedy feature selection based on DNN;
4) \textbf{DeepPink}~\cite{deeppink} combines knockoffs~\cite{knockoff} and Deep Neural Networks to address feature selection problems;
5) \textbf{KnockoffGAN}~\cite{knockoffgan} (short as GAN) utilizes GAN to generate knockoff features that are not limited to Gaussian distribution, enabling feature selection;
6) \textbf{MCDM}~\cite{mcdm} ensemble feature selection as a Multi-Criteria Decision-Making problem, which uses the VIKOR sort algorithm to rank features based on the judgment of multiple feature selection methods;
(B). Embedded methods:
7) \textbf{RFE}~\cite{rfe} recursively deletes the weakest features; 
8) \textbf{LASSO}~\cite{lasso} shrinks the coefficients of useless features to zero by sparsity regularization to select features; 
9) \textbf{LASSONet}~\cite{lassonet} (short as LNet) is a neural network with sparsity to encourage the network to use only a subset of input features; 
(C). Wrapper methods:
10) \textbf{GFS}~\cite{fan2021autogfs} is a group-based feature selection method via interactive reinforcement learning; 
11) \textbf{MARLFS}~\cite{marlfs} uses reinforcement learning to create an agent for each feature to learn a policy to select or deselect the corresponding feature, and treat feature redundancy and downstream task performance as rewards; 
12) \textbf{SARLFS}~\cite{sarlfs} is a simplified version of MARLFS to leverage a single agent to replace multiple agents to decide the selection actions of all features.
To evaluate the necessity of each technical component of \model, we develop two model variants:
i) \textbf{\model$^*$} removes the variational inference component and solely uses the Transformer to create the feature subset embedding space;
ii) \textbf{\model$^{-}$} adopts LSTM~\cite{LSTM} to learn the feature subset embedding space.

\noindent\textbf{Hyperparameters and Reproducibility.}
1) Data Collector: We use the reinforcement data collector to explore 300 epochs to collect feature subset-predictive accuracy data pairs, and randomly shuffle each feature sequence 25 times to augment the training data.
2) Feature Subset Embedding: We map feature tokens to a 64-dimensional embedding, and use a 2-layer network for both encoder and decoder, with a multi-head setting of 8 and a feed-forward layer dimension of 256. The latent dimension of the VAE is set to 64. The estimator consists of a 2-layer feed-forward network, with each layer having a dimension of 200. The values of $\alpha$, $\beta$, and $\gamma$ are 0.8, 0.2, and 0.001, respectively. We set the batch size as 1024, the training epochs as 100, and the learning rate as 0.0001. 
3) Optimal Embedding Search and Reconstruction: We use the top 25 feature sets to search for the feature subsets and keep the optimal feature subset. 

\noindent\textbf{Environmental Settings.}
All experiments are conducted on the Ubuntu 22.04.3 LTS operating system, Intel(R) Core(TM) i9-13900KF CPU@ 3GHz, and 1 way RTX 4090 and 32GB of RAM, with the Python 3.11.4 and PyTorch 2.0.1.

\setlength{\tabcolsep}{0.5mm}{
\begin{table*}[tb]
\centering
\small
\caption{Overall Performance. The best and the second-best results are highlighted by \textbf{bold} and \underline{underlined} fonts respectively. We evaluate classification (C) and regression (R) tasks in terms of F1-score and 1-RAE respectively. 
The higher the value is, the better the feature space quality is.
The bold percentage reflects the improvements of \model\ compared with the best baseline model.}
\begin{tabular}{@{}ccccccccccccccc@{}}
\toprule
\toprule
Dataset            & Original & K-Best & mRMR                               & DNP   & DeepPink & GAN                        & MCDM                               & RFE                                & LASSO                              & LNet                           & GFS                                & MARLFS                             & SARLFS                             & \textbf{\model} \\ \midrule
SpectF             & 75.96    & 78.21  & 78.21                              & 80.80 & 75.01    & 79.16                              & 80.36                              & \underline{80.80} & 79.16                              & 75.96                              & 75.01                              & 75.01                              & 79.16                              & \textbf{84.58(+4.68\%)}        \\
SVMGuide3          & 77.81    & 76.84  & 76.84                              & 77.12 & 76.55    & 77.91                              & 76.66                              & 78.07                              & 77.91                              & 76.44                              & \underline{83.12} & 76.84                              & 76.22                              & \textbf{85.02(+2.29\%)}        \\
German Credit      & 64.88    & 66.79  & 66.79                              & 68.43 & 64.88    & 66.31                              & \underline{70.85} & 64.86                              & 66.4                               & 63.97                              & 67.54                              & 66.31                              & 63.12                              & \textbf{73.50(+3.74\%)}        \\
UCI Credit         & 80.19    & 80.59  & \underline{80.59} & 79.94 & 80.43    & \underline{80.59} & 74.46                              & 80.28                              & 77.94                              & 80.05                              & 79.96                              & 80.24                              & 80.05                              & \textbf{81.21(+0.77\%)}        \\
SpamBase           & 92.68    & 92.02  & 92.34                              & 91.79 & 92.68    & 92.34                              & 88.95                              & 91.68                              & 91.81                              & 91.67                              & 92.25                              & \underline{92.35} & 90.94                              & \textbf{93.53(+1.28\%)}        \\
Ap\_omentum & 66.19    & \underline{84.49}  & \underline{84.49}                              & 82.03 & 82.03    & \underline{84.49}                              & \underline{84.49} & 84.49                              & 82.03                              & 83.02                              & 82.03                              & \underline{84.49} & \underline{84.49} & \textbf{86.52(+2.40\%)}        \\
Ionosphere         & 92.85    & 91.32  & 94.27                              & 94.12 & 92.85    & 94.27                              & 88.64                              & \underline{95.69} & 88.17                              & 88.38                              & 91.34                              & 89.92                              & 88.51                              & \textbf{97.13(+1.50\%)}        \\
Activity           & 96.17    & 96.07  & 95.92                              & 95.87 & 96.12    & \underline{96.17} & 96.12                              & 95.87                              & 95.92                              & \underline{96.17} & 96.12                              & 95.87                              & 95.87                              & \textbf{97.33(+1.21\%)}        \\
Mice-Protein       & 74.99    & 77.32  & 78.68                              & 77.29 & 77.47    & 78.68                              & 78.69                              & 77.29                              & \underline{78.71} & 76.4                               & 77.35                              & 76.4                               & 74.53                              & \textbf{81.96(+4.13\%)}        \\ \midrule
Openml-586         & 54.95    & 57.68  & 57.64                              & 60.74 & 58.47    & 60.74                              & 57.95                              & 58.1                               & 60.67                              & 58.28                              & \underline{62.27} & 58.27                              & 56.98                              & \textbf{63.99(+2.76\%)}        \\
Openml-589         & 50.95    & 57.17  & 57.17                              & 54.68 & 57.42    & 57.17                              & 55.43                              & 54.25                              & \underline{58.74} & 57.55                              & 44.72                              & 57.39                              & 53.48                              & \textbf{61.13(+4.07\%)}        \\
Openml-607         & 51.73    & 54.64  & 55.17                              & 55.14 & 55.68    & 57.88                              & 55.56                              & 54.39                              & \underline{58.10} & 55.38                              & 45.7                               & 54.99                              & 53.28                              & \textbf{62.72(+7.95\%)}        \\
Openml-616         & 15.63    & 26.95  & 25.45                              & 25.93 & 26.74    & 28.56                              & 22.92                              & 24.08                              & \underline{28.98} & 25.98                              & 22.93                              & 26.29                              & 23.06                              & \textbf{33.85(+16.8\%)}        \\
Openml-618         & 46.89    & 51.79  & 51.08                              & 51.73 & 51.46    & \underline{52.40} & 50.9                               & 50.64                              & 47.41                              & 51.11                              & \underline{52.40} & 51.87                              & 48.54                              & \textbf{55.91(+6.69\%)}        \\
Openml-620         & 51.01    & 55.03  & 55.03                              & 55.66 & 55.66    & 55.94                              & 55.66                              & 53.96                              & 57.99                              & 55.94                              & \underline{58.99} & 55.42                              & 53.98                              & \textbf{62.58(+6.09\%)}        \\
Openml-637         & 14.95    & 21.06  & 20.49                              & 20.45 & 20.47    & 21.12                              & 22.16                              & 17.82                              & 26.02                              & 19.43                              & \underline{39.12} & 20.75                              & 19.45                              & \textbf{42.18(+7.82\%)}        \\ \bottomrule \bottomrule
\end{tabular}
\label{exp:overall_performance}
\end{table*}}

\subsection{Overall Performance.}
In this experiment, we evaluate the performance of {\model} and baseline algorithms for feature selection on 16 datasets in terms of F1-score or 1-RAE. 
Table~\ref{exp:overall_performance} shows the comparison results.
We can find that {\model} consistently surpasses other baseline models across all datasets, achieving an average performance improvement of 3\% over the second-best baseline model.
The underlying driver of this observation is that \model\ can compress the feature learning knowledge into a large embedding space.
Such a compression facilitates a more effective search for the optimal feature selection result.
Moreover, another interesting observation is that the algorithm ranking second-best varies across different datasets.
A possible reason for the observation is that traditional feature selection methods are designed based on varying criteria, resulting in a limited generalization capability across different scenarios.
In summary, this experiment shows the effectiveness of VTFS in feature selection, underscoring the great potential of generative AI in this domain.

\begin{figure*}[tb]
	\centering
	\subfigure[SpectF]{\label{exp:vae:spect}\includegraphics[width=0.237\textwidth]{{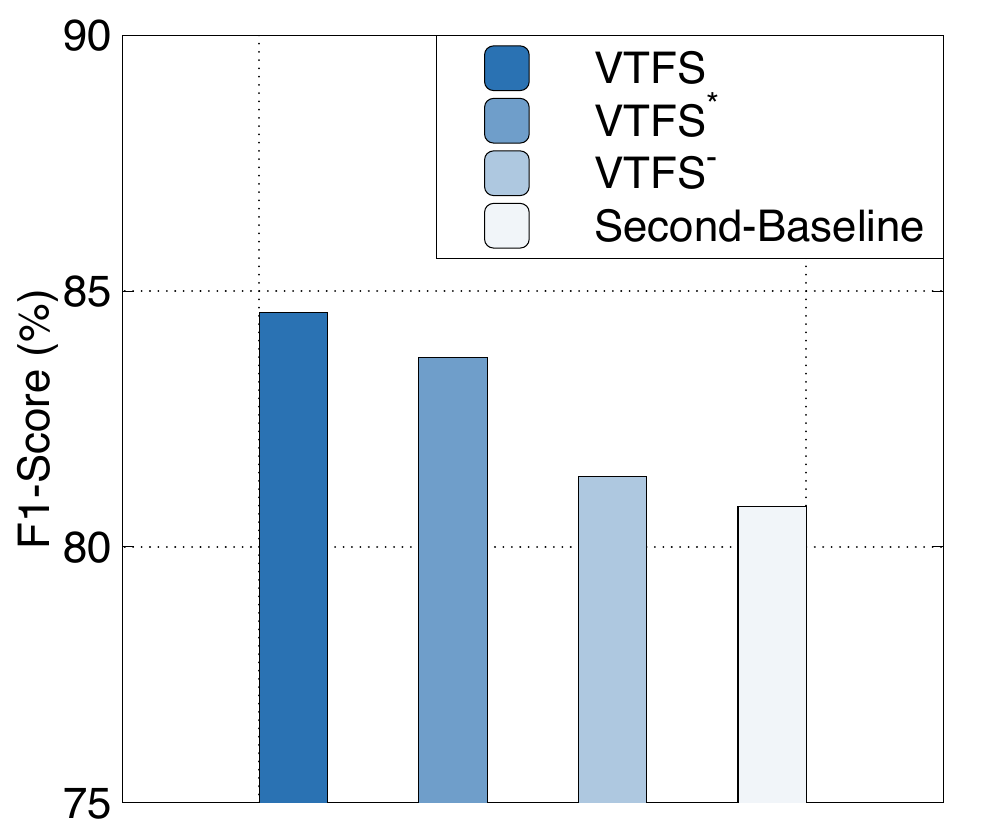}}}
	\subfigure[SVMGuide3]{\label{exp:vae:svm}\includegraphics[width=0.237\textwidth]{{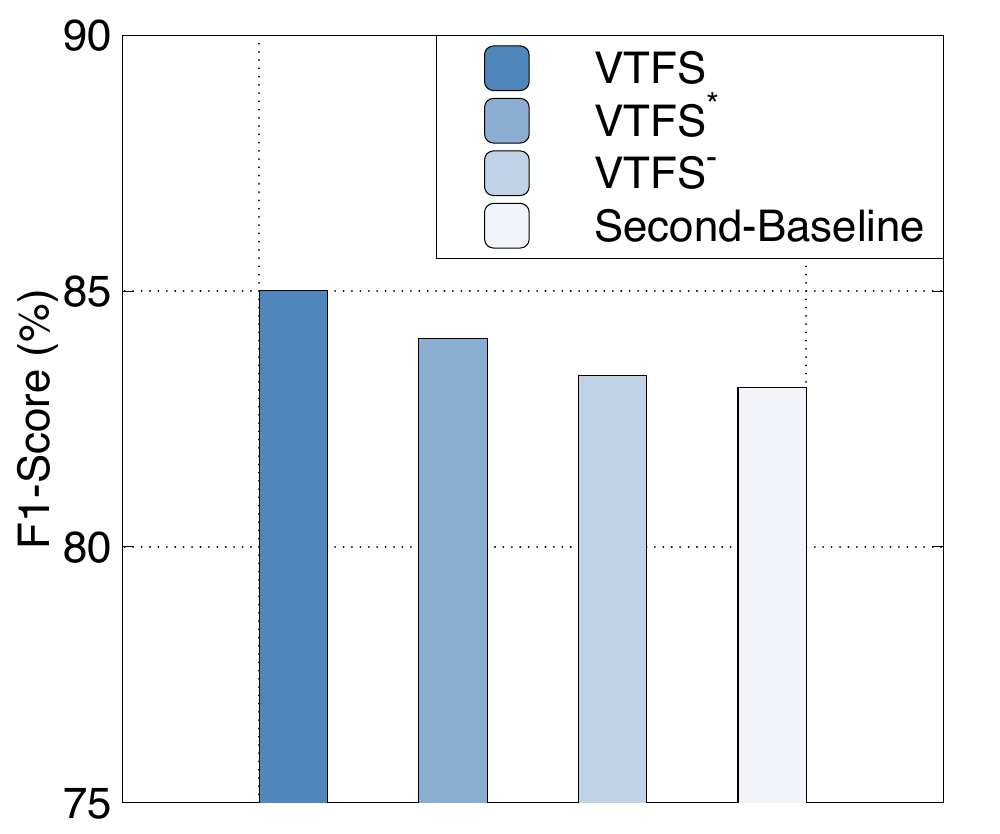}}}
         \subfigure[German Credit]{\label{exp:vae:german}\includegraphics[width=0.237\textwidth]{{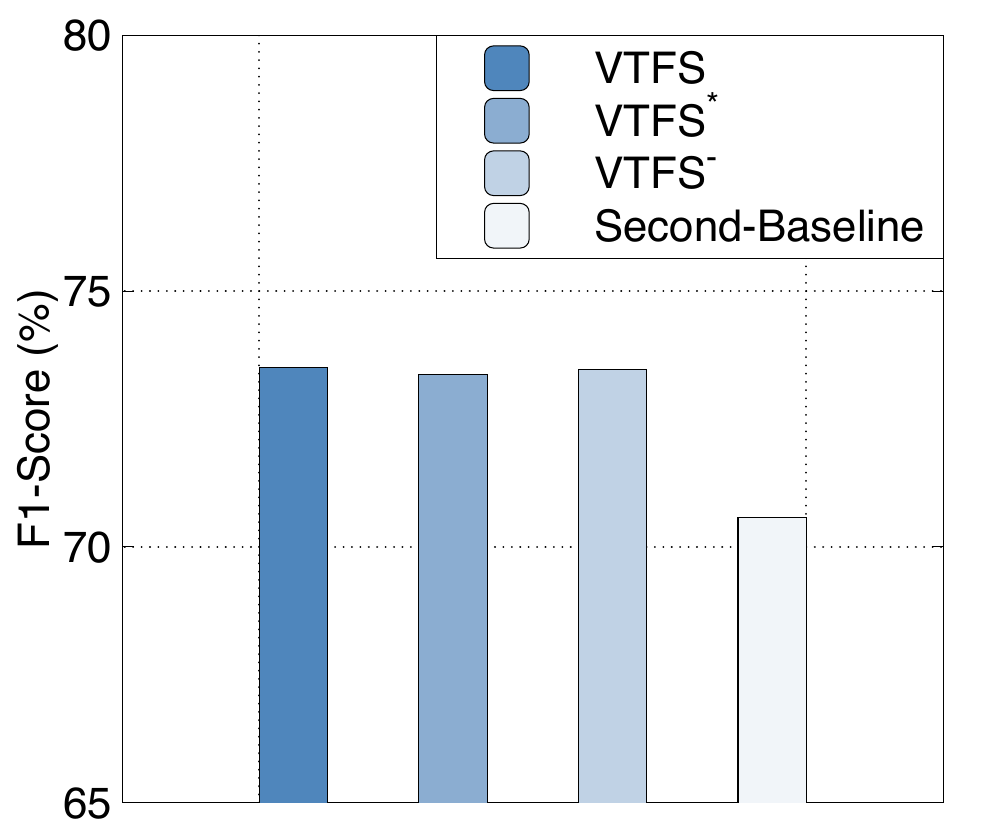}}}
         \subfigure[UCI Credit]{\label{exp:vae:UCI}\includegraphics[width=0.237\textwidth]{{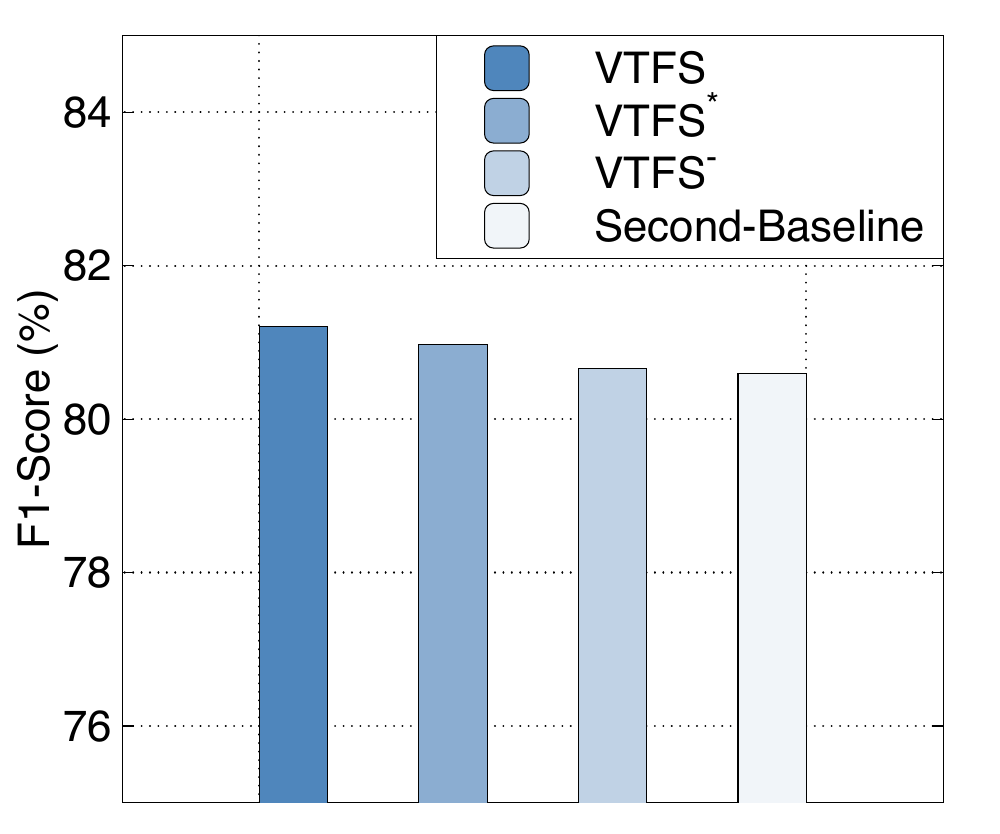}}}
         \subfigure[SpamBase]{\label{exp:vae:spam}\includegraphics[width=0.237\textwidth]{{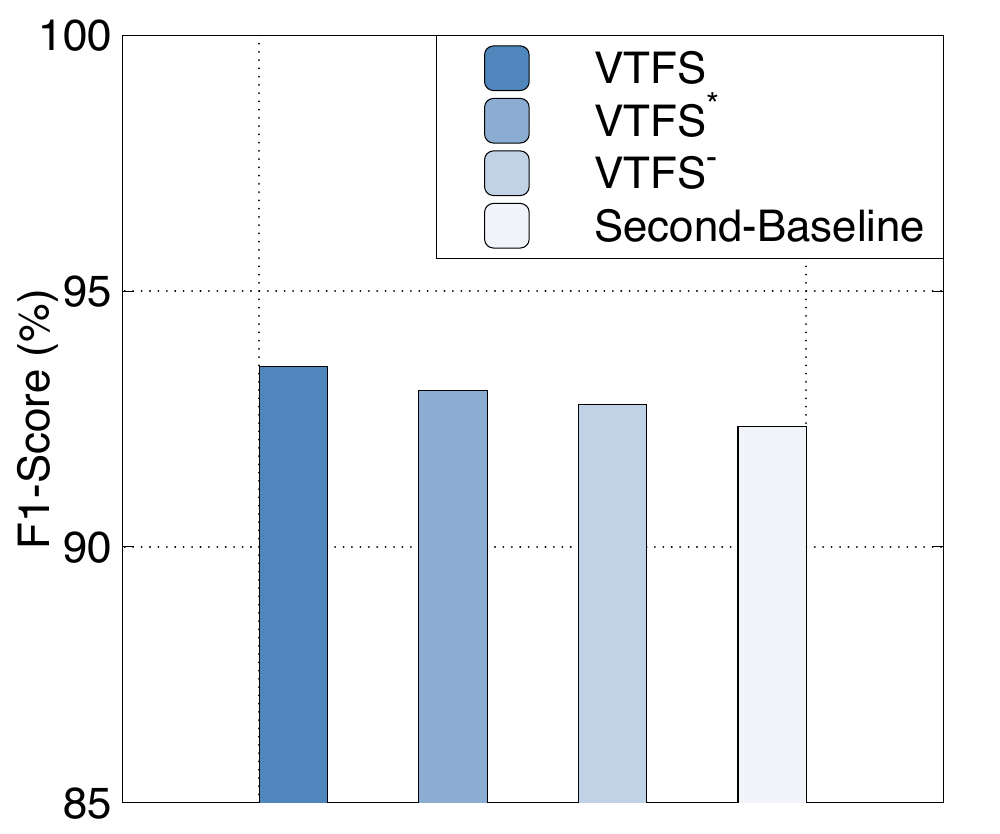}}}
         \subfigure[Ap Omentum]{\label{exp:vae:ap}\includegraphics[width=0.237\textwidth]{{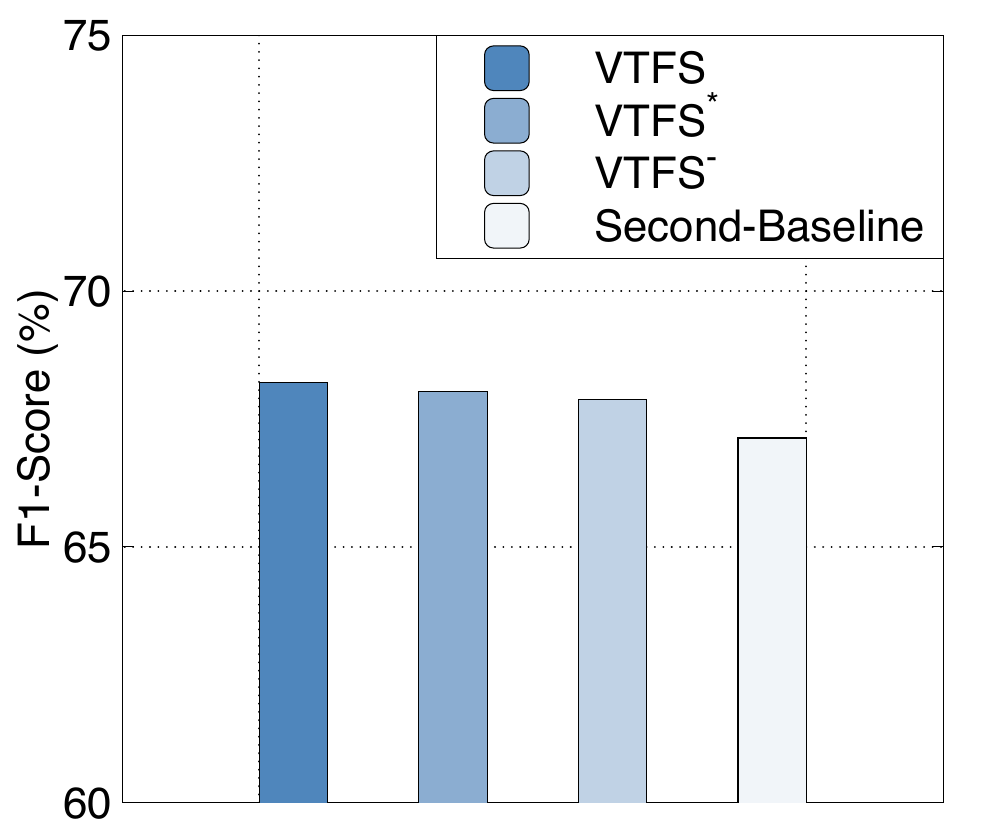}}}
         \subfigure[IonoSphere]{\label{exp:vae:iono}\includegraphics[width=0.237\textwidth]{{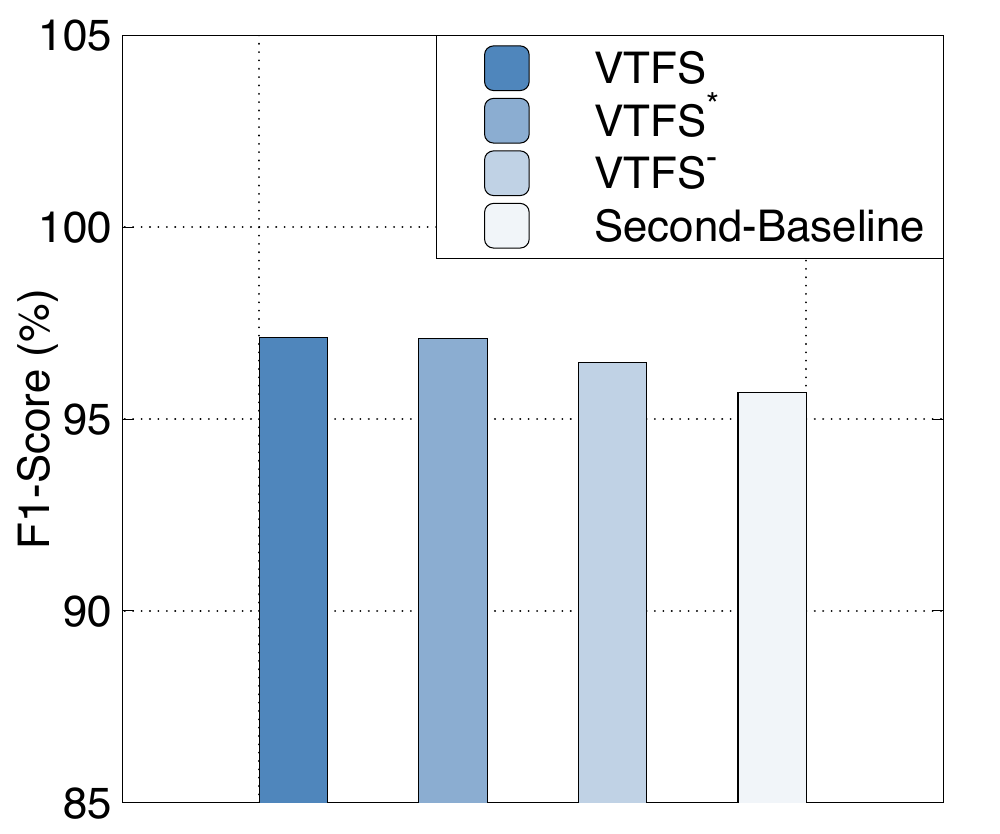}}}
         \subfigure[Acticity]
         {\label{exp:vae:activity}\includegraphics[width=0.237\textwidth]{{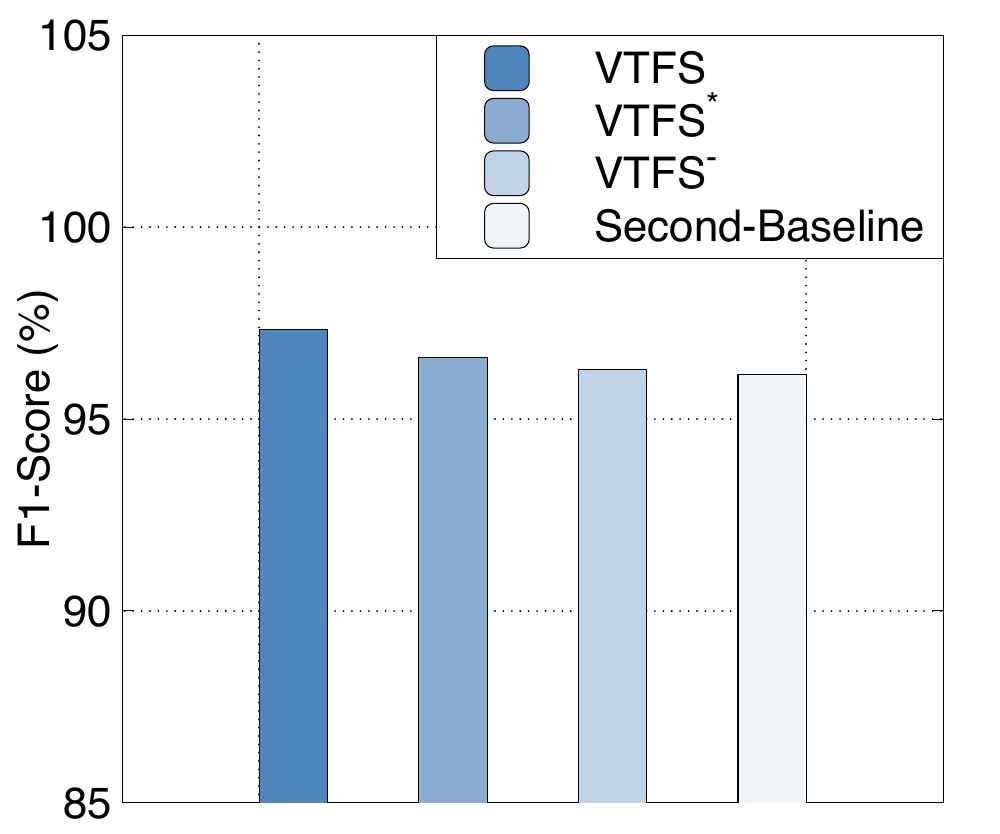}}}
         \subfigure[Mice-Protein]{\label{exp:vae:mice}\includegraphics[width=0.237\textwidth]{{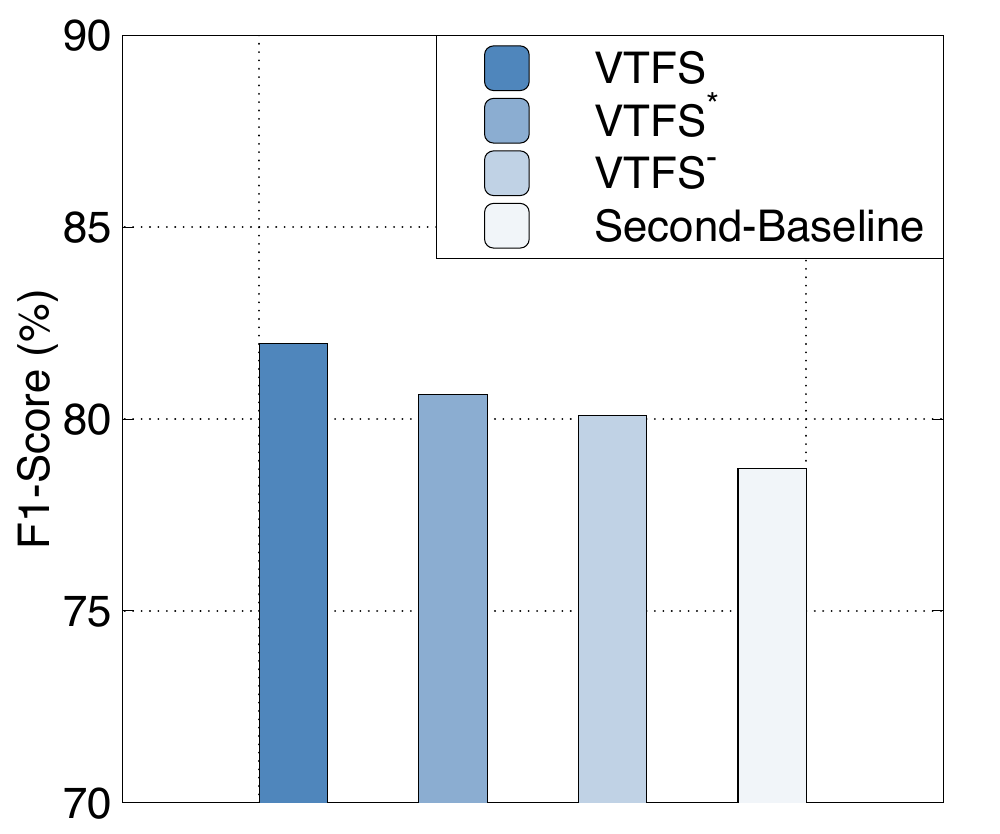}}}
         \subfigure[openml-586]{\label{exp:vae:reg-586}\includegraphics[width=0.237\textwidth]{{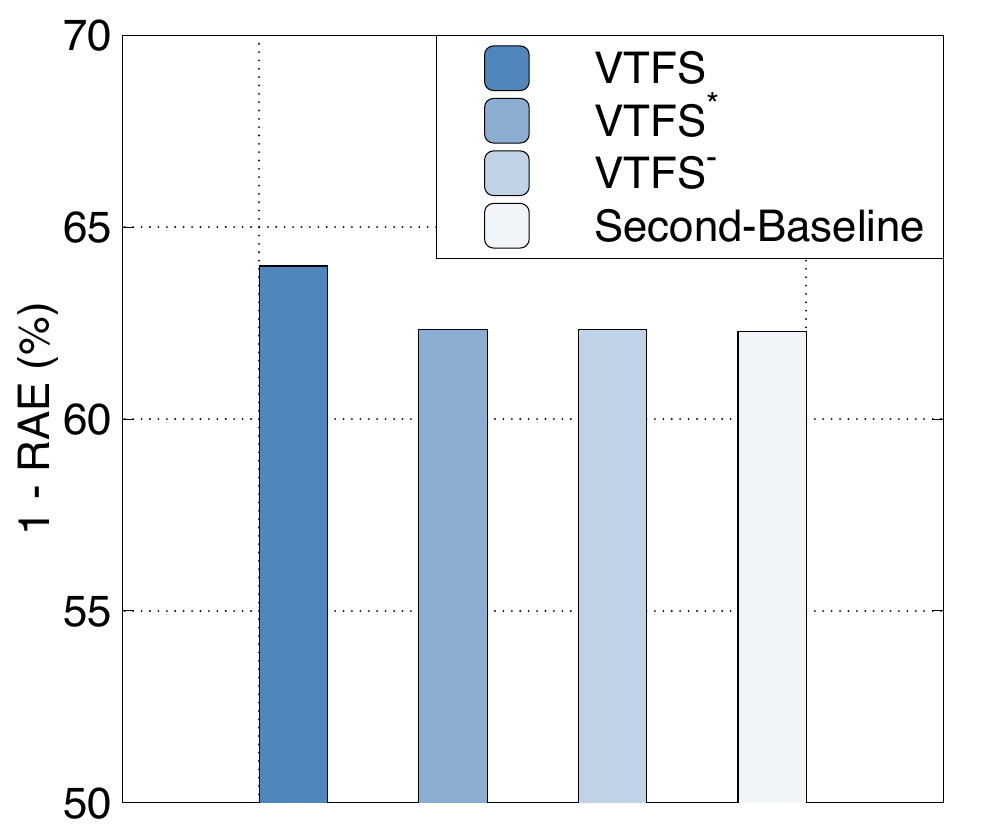}}}
         \subfigure[openml-589]{\label{exp:vae:reg-589}\includegraphics[width=0.237\textwidth]{{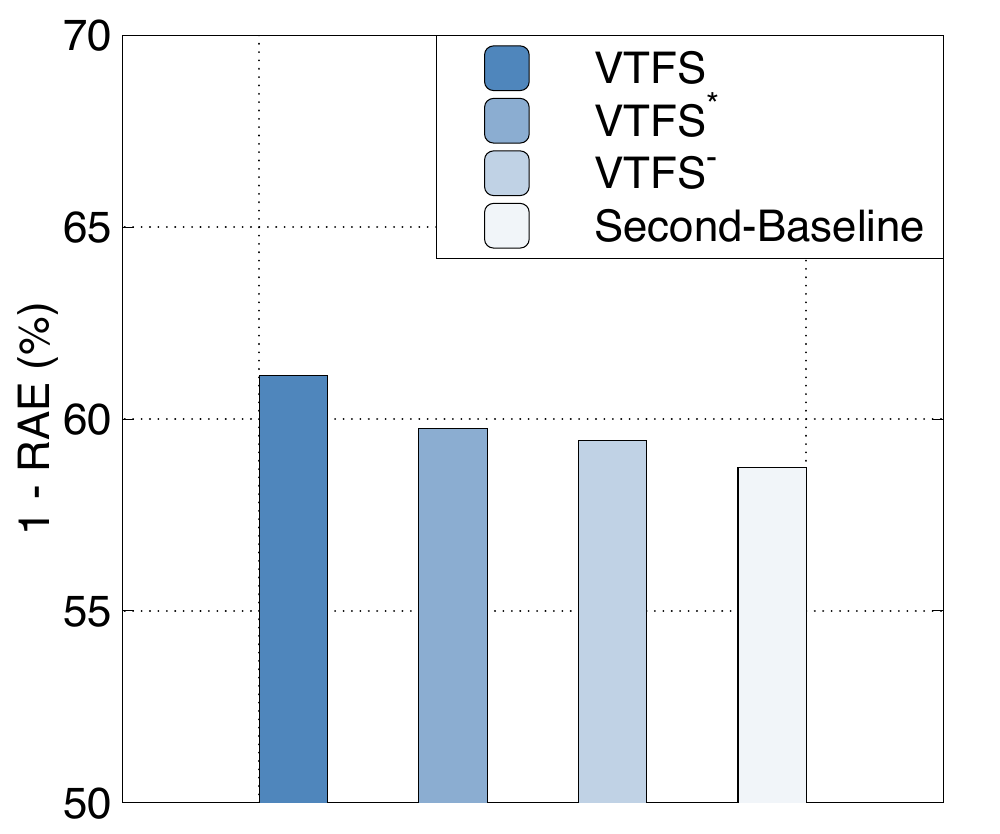}}}
	\subfigure[openml-607]{\label{exp:vae:reg-607}\includegraphics[width=0.237\textwidth]{{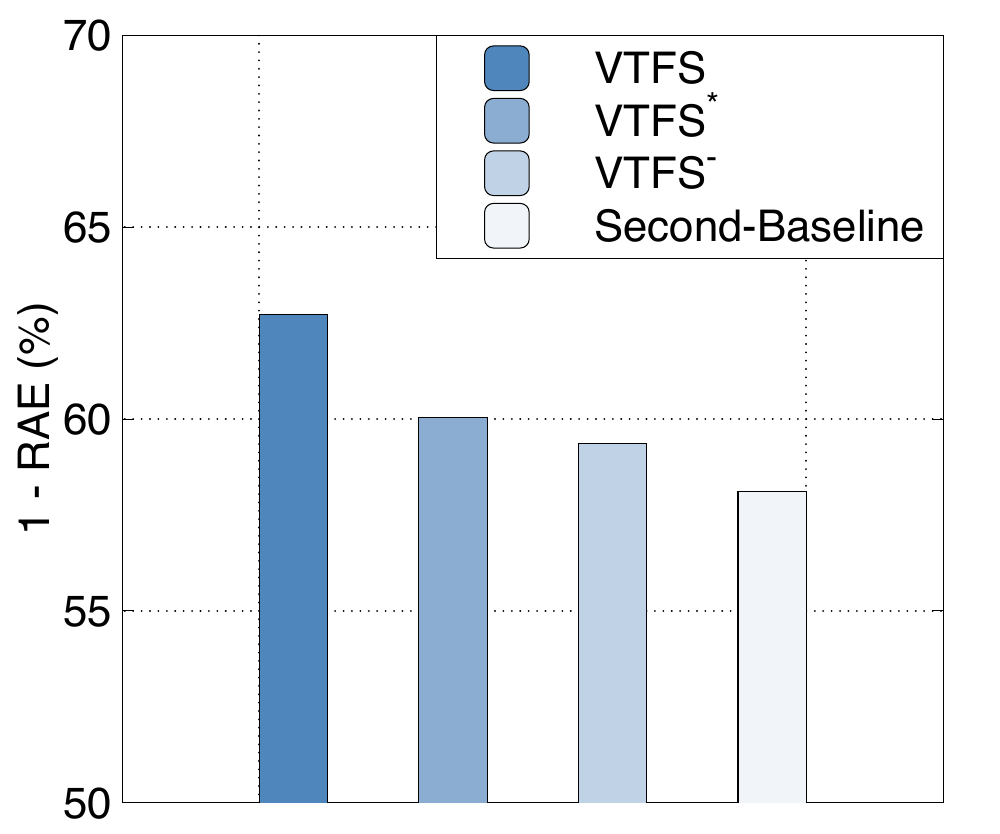}}}
         \subfigure[openml-616]{\label{exp:vae:reg-616}\includegraphics[width=0.237\textwidth]{{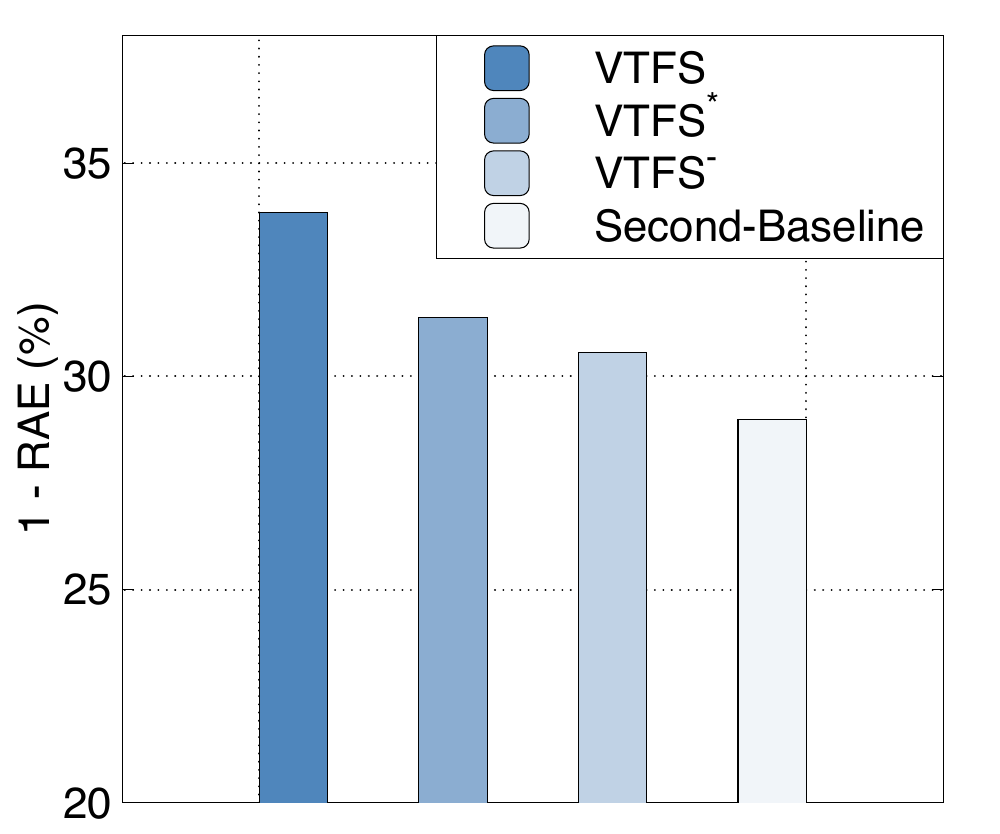}}}
         \subfigure[openml-618]{\label{exp:vae:reg-618}\includegraphics[width=0.237\textwidth]{{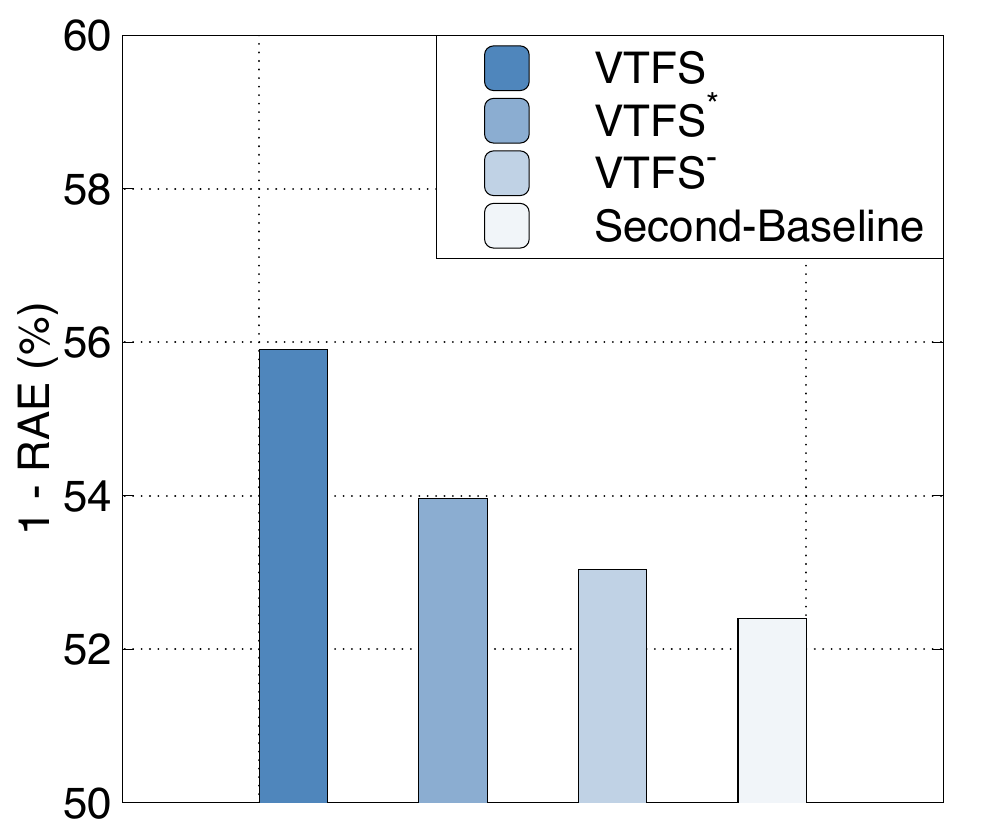}}}
	\subfigure[openml-620]{\label{exp:vae:reg-620}\includegraphics[width=0.237\textwidth]{{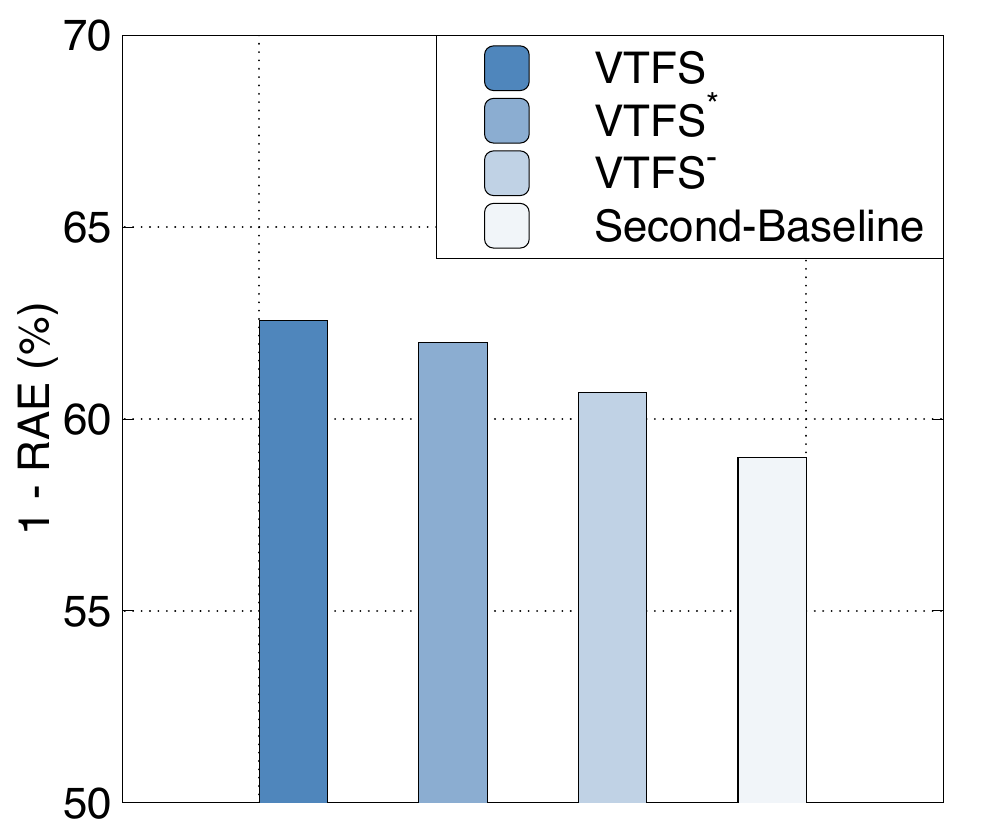}}}
         \subfigure[openml-637]{\label{exp:vae:reg-637}\includegraphics[width=0.237\textwidth]{{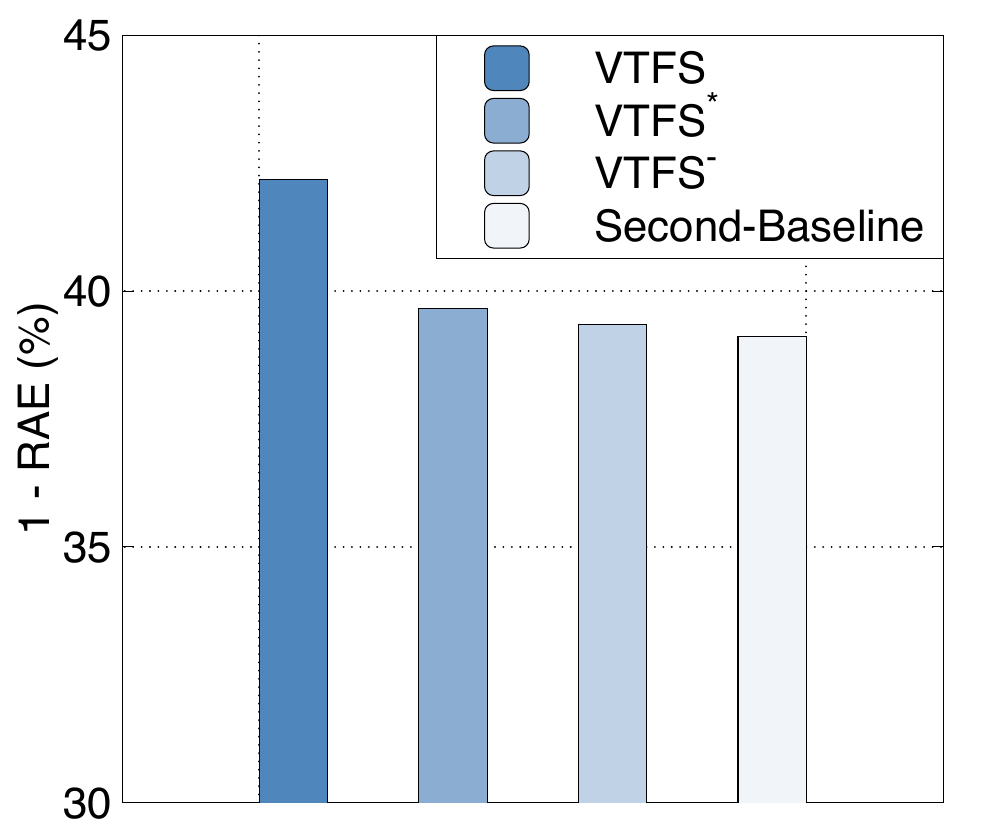}}}
	\caption{Analysis of the impact of different feature subset embedding modules on feature selection.}
	\label{exp:ablation_seq}
 \vspace{-0.3cm}
\end{figure*}

\subsection{Study of the influence of variational transformer for continuous space construction.} 
One of the important novelties of \model\ involves a sequential model to embed feature learning knowledge into an embedding space.
To analyze the influence of the selection of the sequential model, we develop two model variants: 1) \model$^-$, which employs an LSTM model as the backbone of the sequential model; 2) \model$^*$, which removes the variational inference component and exclusively uses a transformer model.
Figure~\ref{exp:ablation_seq} shows the comparison results in terms of F1-score and 1-RAE for classification and regression tasks respectively.
We can find that \model\ outperforms \model$^*$ with a great performance gap across all datasets.
The underlying driver for this observation is that the variational inference component in \model\ enhances the smoothness of the learned feature subset embedding space.
This smoothness facilitates a more effective search for optimal feature selection results.
Additionally, another interesting observation is that \model$^*$ surpasses \model$^-$ across all datasets in both classification and regression tasks.
A potential reason for this observation is that the transformer architecture, compared to LSTM, is more adept at capturing complex correlations between different feature combinations and their impact on downstream machine learning task performance.
Moreover, it is noticed that even when solely employing LSTM, \model$^-$ still outperforms the second-best baseline algorithm across various datasets.
This observation underscores the success and effectiveness of the generative AI perspective of \model.
In conclusion, this experiment indicates the necessity of each technical component of \model.

\begin{figure}
	\centering
	\subfigure[German Credit]{\includegraphics[width=0.237\textwidth]{{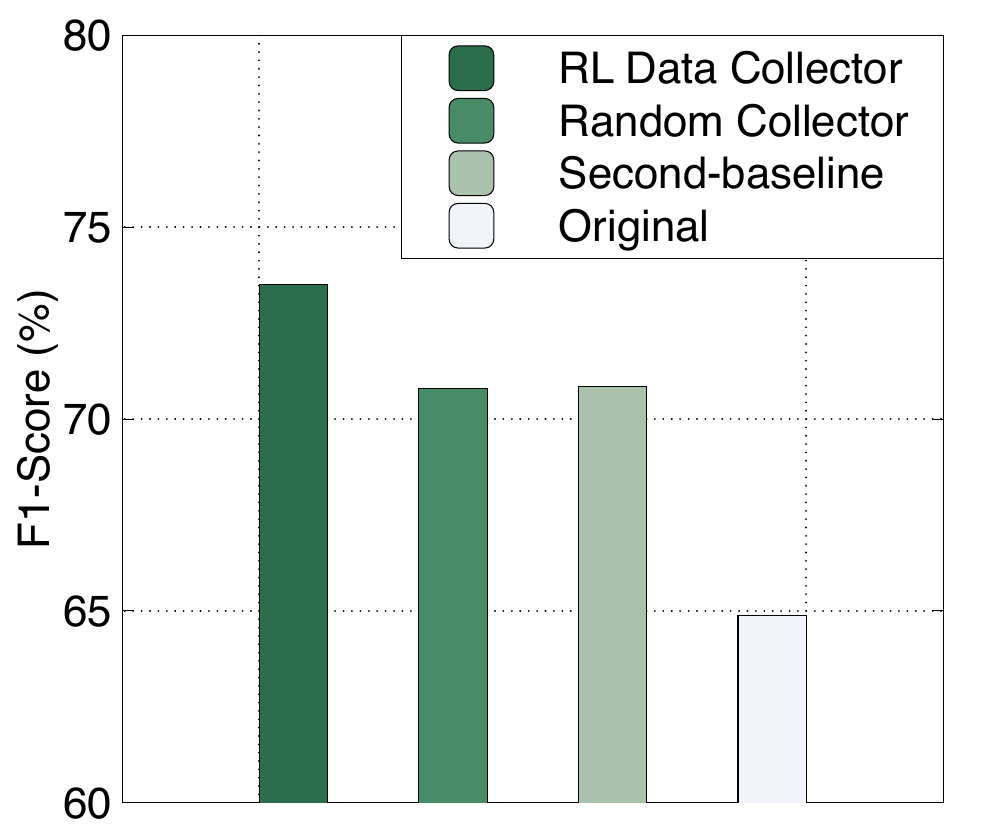}}}
	\subfigure[Activity]{\includegraphics[width=0.237\textwidth]{{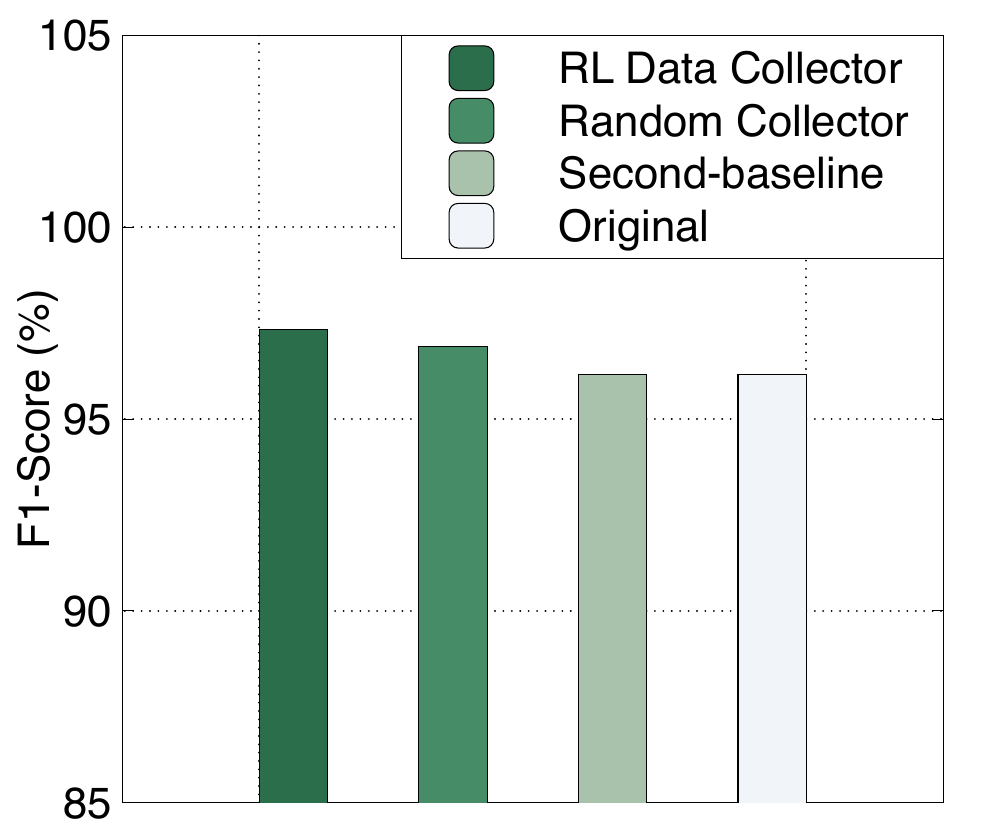}}}
         \subfigure[openml-586]{\includegraphics[width=0.237\textwidth]{{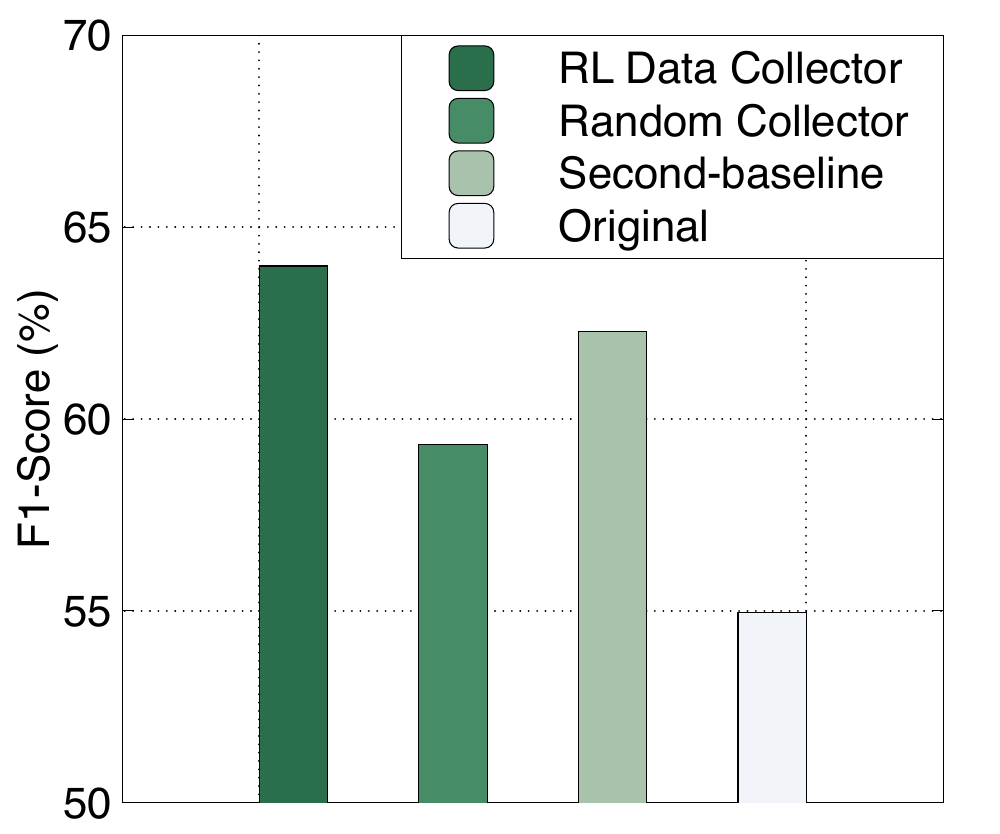}}}
         \subfigure[openml-589]{\includegraphics[width=0.237\textwidth]{{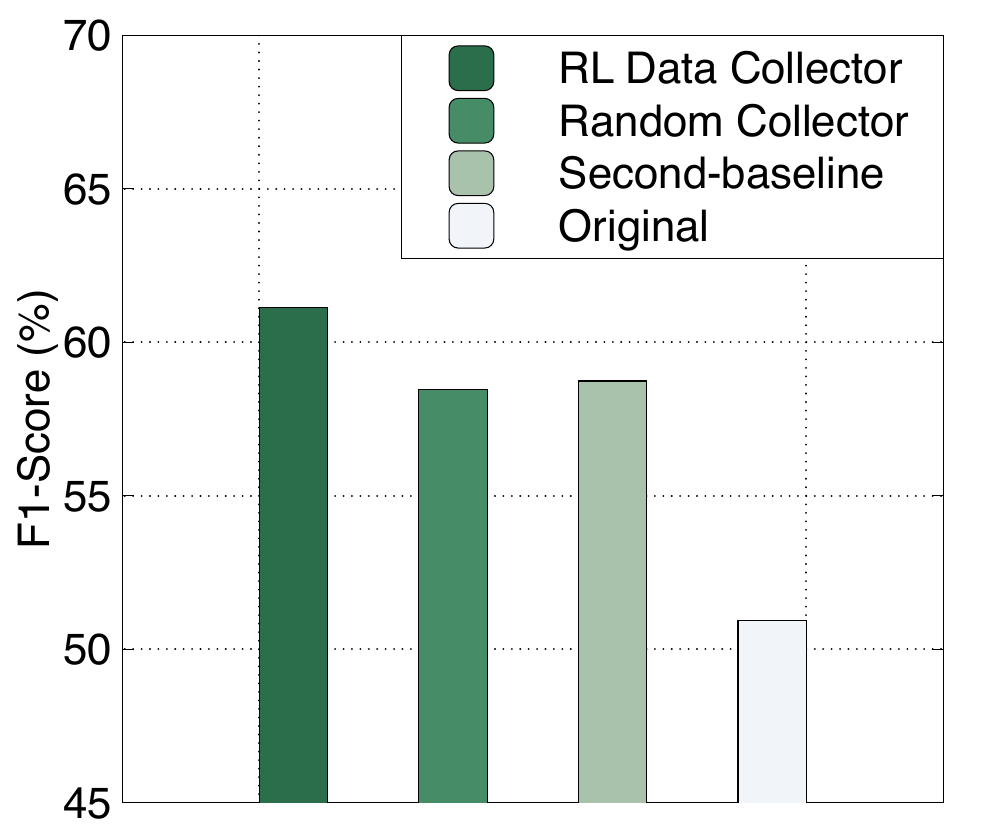}}}
        \caption{Analysis of the impact of data collector on selecting the effective feature subset.}
	\label{exp:rl_data_collector}
 \vspace{-0.3cm}
\end{figure}
\begin{figure}
	\centering
	\subfigure[German Credit]{\includegraphics[width=0.237\textwidth]{{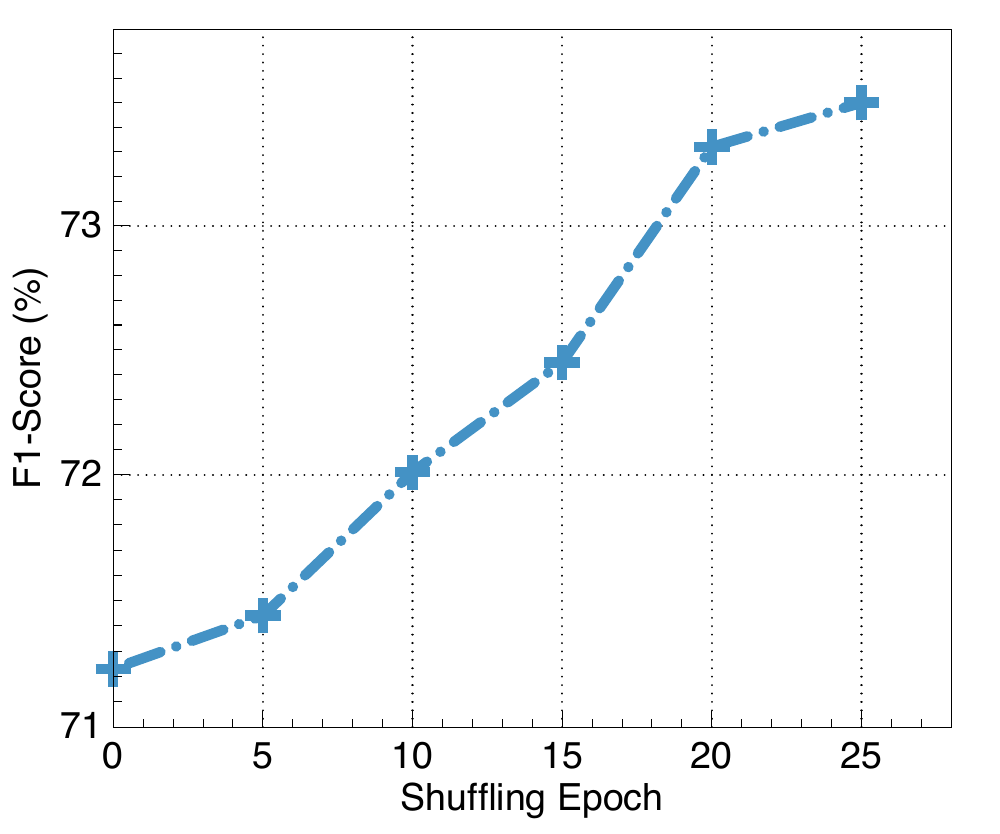}}}
	\subfigure[Activity]{\includegraphics[width=0.237\textwidth]{{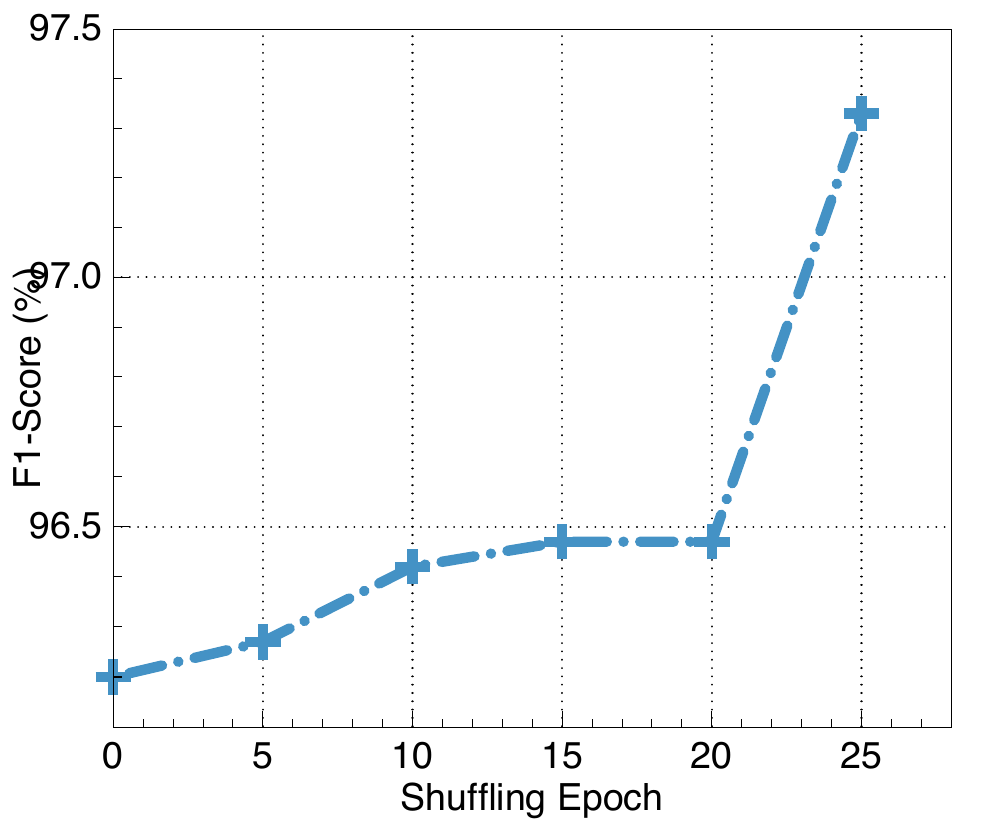}}}
         \subfigure[openml-586]{\includegraphics[width=0.237\textwidth]{{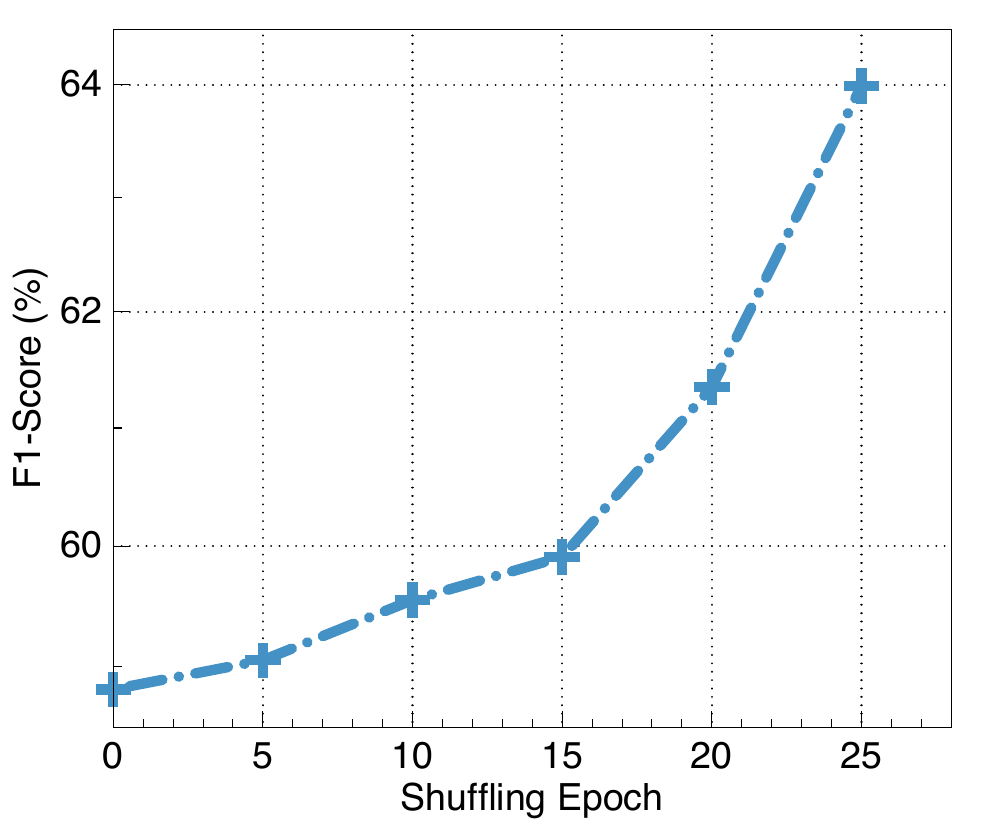}}}
         \subfigure[openml-589]{\includegraphics[width=0.237\textwidth]{{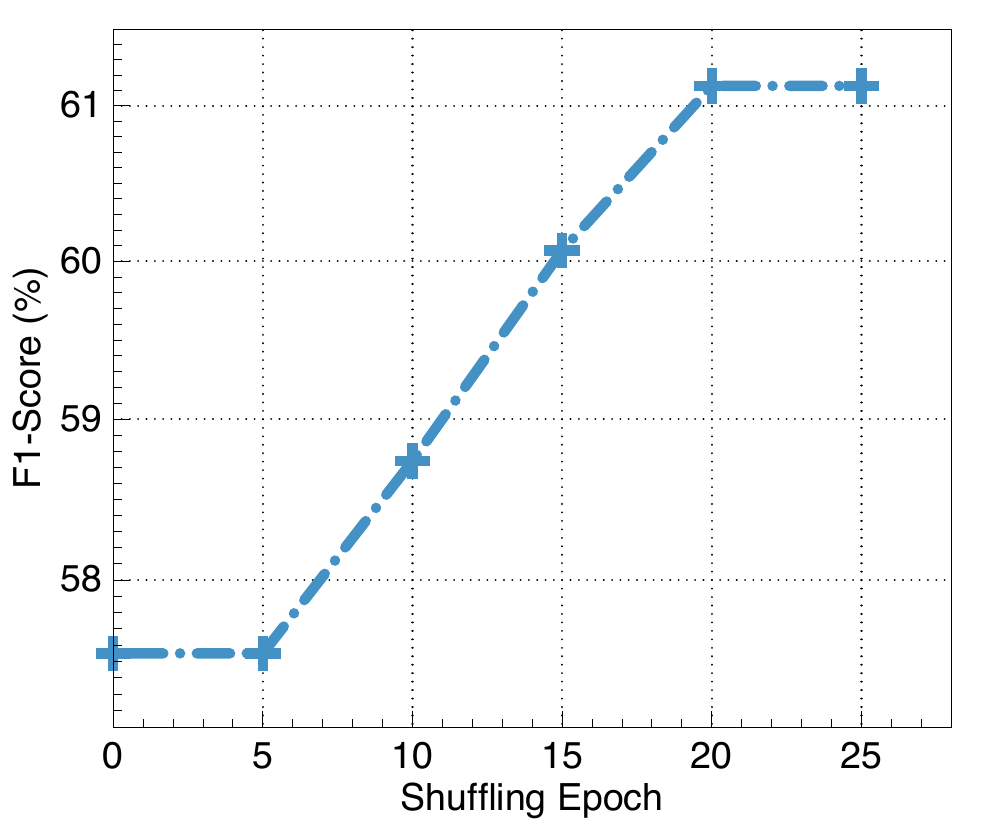}}}
        \caption{Analysis of the impact of data augmentation on selecting the effective feature subset.}
	\label{exp:data_augmentation}
 \vspace{-0.3cm}
\end{figure}

\subsection{Study of the impact of RL-based data collector.}
In \model, we emphasize the capability of the RL-based data collector to gather higher-quality and more diverse training data, thereby facilitating the construction of a better embedding space. To assess the impact of the RL-based data collector, we established three control groups on four datasets: 1) randomly collecting training data samples to construct the feature subset embedding space and generate the feature subset; 2) using the second best baseline of each data set to obtain the feature subset; 3) directly using original feature set for prediction. \textbf{Figure~\ref{exp:rl_data_collector}} shows that the training data collected by the RL-data collector can help identify a feature subset superior to all control groups. The underlying driver is that the RL-based data collector can produce higher-quality and diverse data, contributing to the creation of a more effective embedding space. This enhanced embedding space facilitates the identification of the best feature subset based on the gradient search method. Another observation is when constructing the embedding space using randomly collected data and subsequently searching for the optimal feature subset, the performance in the downstream ML task significantly improves compared to the original feature set but in three cases worse than the second-best baselines. This suggests that \model\ can learn feature subset knowledge, thereby identifying an effective feature subset to improve downstream performance. However, collecting diverse training data is necessary and important, which makes the embedding space more distinguishable to identify superior feature selection outcomes.
In summary, this experiment demonstrates that the RL-based data collector is an indispensable component to maintain the excellent feature selection performance of \model.

\setlength{\tabcolsep}{3.1mm}{
\begin{table}
\centering
\small
\caption{Time and Space complexity analysis in terms of the feature size, running time, and parameter size.}
\begin{tabular}{ccccccc}
\toprule \toprule
                   & \#Features & \#Samples & \begin{tabular}[c]{@{}c@{}}Data Collect\\ 300 epochs\end{tabular} & \begin{tabular}[c]{@{}c@{}}Parameter \\ Size\end{tabular} & \begin{tabular}[c]{@{}c@{}}Training Time\\ 100 epochs\end{tabular} & \begin{tabular}[c]{@{}c@{}}Inference \\ Time\end{tabular} \\ \hline
SpectF             & 44         & 267       & 66.15                                                             & 0.387231MB                                                & 67.68                                                              & 0.29                                                       \\
SVMGuide3          & 21         & 1243      & 140.45                                                            & 0.382792MB                                                & 55.15                                                              & 0.13                                                       \\
German Credit      & 24         & 1001      & 102.18                                                            & 0.383371MB                                                & 65.93                                                              & 0.12                                                       \\
UCI Credit         & 25         & 30000     & 2710.51                                                           & 0.383371MB                                                & 63.67                                                              & 0.12                                                       \\
SpamBase           & 57         & 4601      & 390.88                                                            & 0.38974MB                                                 & 66.95                                                              & 0.40                                                       \\
Ap\_omentum & 10936      & 275       & 10315.71                                                          & 2.489387MB                                                & 1118.52                                                            & 22.08                                                      \\
Ionosphere         & 34         & 351       & 67.96                                                             & 0.385301MB                                                & 61.14                                                              & 0.16                                                       \\
Activity           & 561        & 10299     & 9052.31                                                           & 0.487012MB                                                & 762.85                                                             & 4.53                                                       \\
Mice-Protein       & 77         & 1080      & 634.77                                                            & 0.3936MB                                                  & 68.35                                                              & 0.51                                                       \\
Openml-586         & 25         & 1000      & 225.42                                                            & 0.383564MB                                                & 61.74                                                              & 0.12                                                       \\
Openml-589         & 25         & 1000      & 209.02                                                            & 0.383564MB                                                & 61.53                                                              & 0.12                                                       \\
Openml-607         & 50         & 1000      & 326.04                                                            & 0.388389MB                                                & 70.12                                                              & 0.32                                                       \\
Openml-616         & 50         & 500       & 165.69                                                            & 0.388389MB                                                & 68.17                                                              & 0.32                                                       \\
Openml-618         & 50         & 1000      & 341.57                                                            & 0.388389MB                                                & 70.56                                                              & 0.32                                                       \\
Openml-620         & 25         & 1000      & 209.22                                                            & 0.383564MB                                                & 62.89                                                              & 0.12                                                       \\
Openml-637         & 50         & 500       & 164.20                                                            & 0.388389MB                                                & 68.87                                                              & 0.32                                                       \\ \bottomrule \bottomrule
\end{tabular}
\label{exp:complexity_check}
\end{table}}
\begin{figure}
	\centering
	\subfigure[Training Time]{\includegraphics[width=0.237\textwidth]{{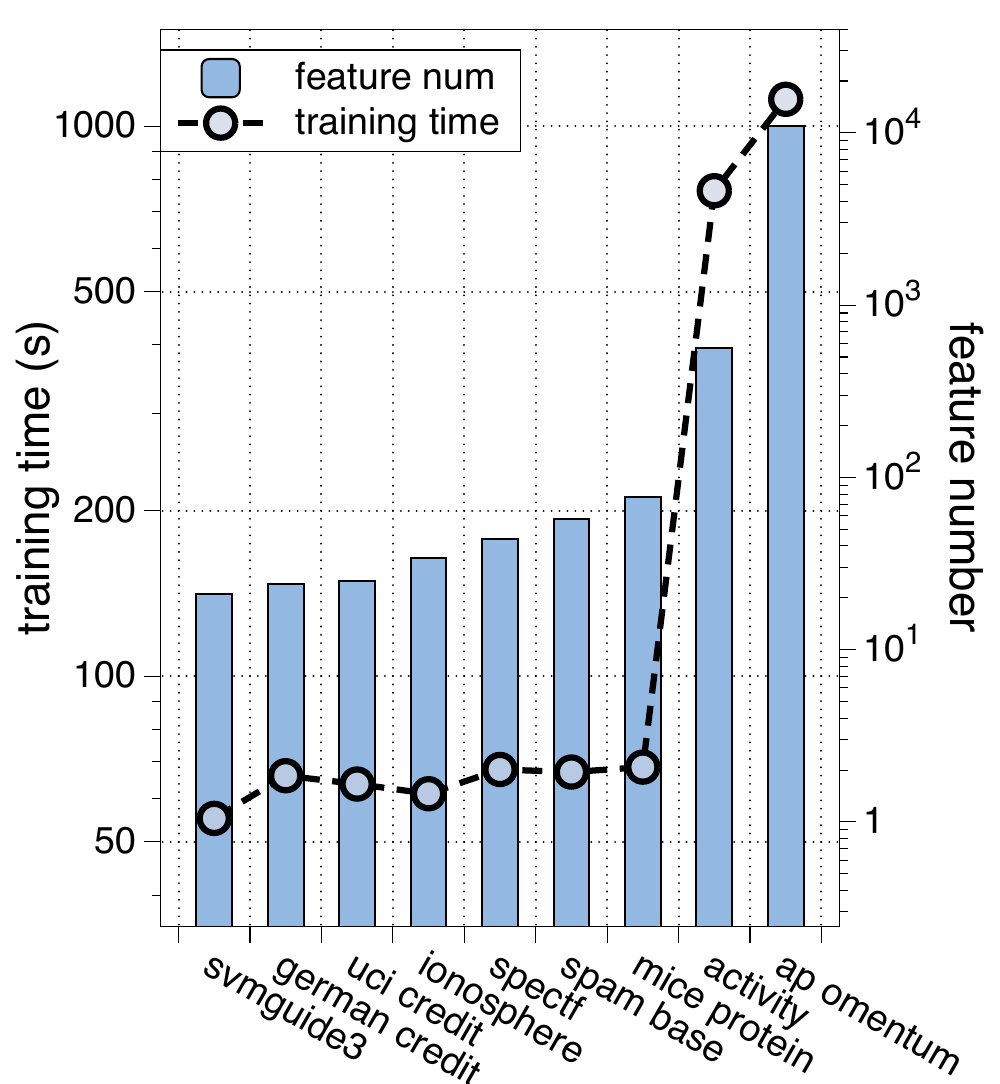}}}
	\subfigure[Inference Time]{\includegraphics[width=0.237\textwidth]{{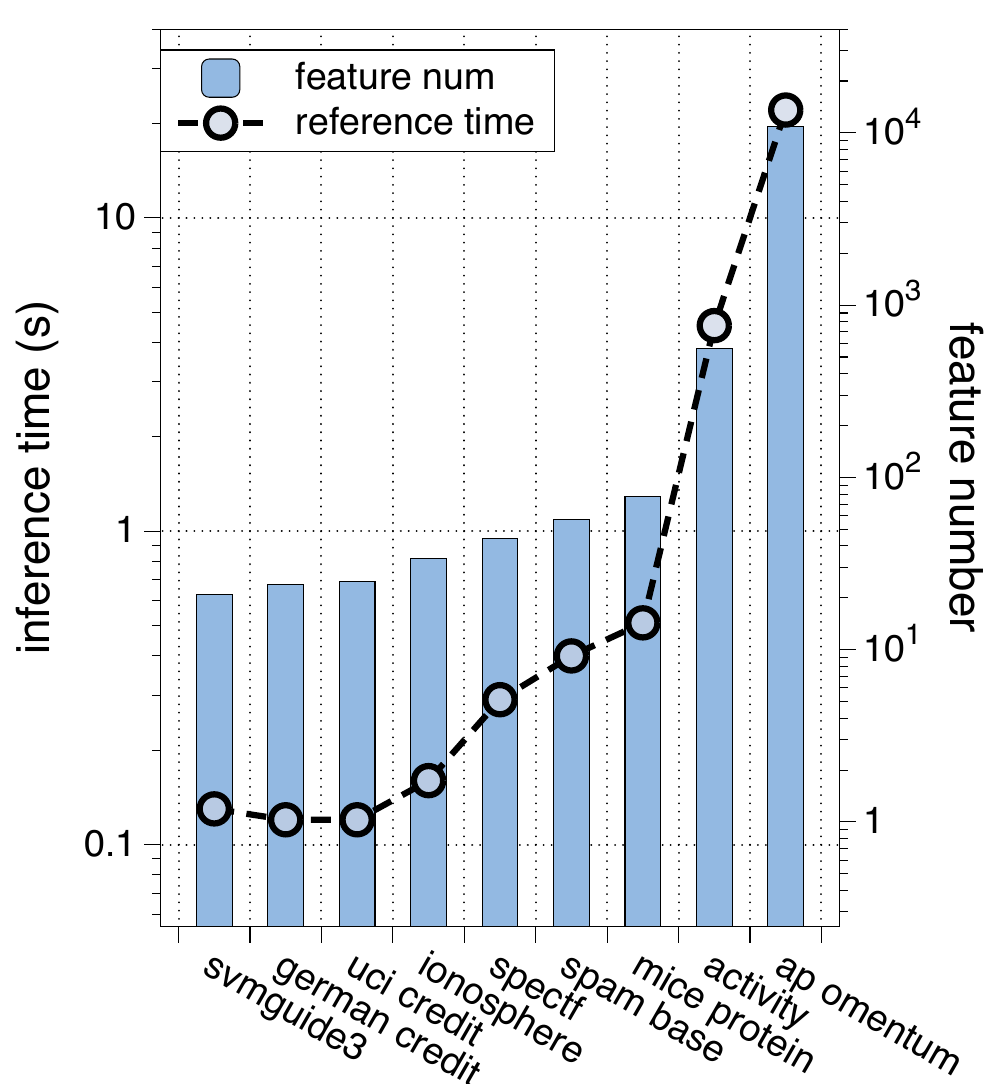}}}
         \subfigure[Parameter Size]{\includegraphics[width=0.237\textwidth]{{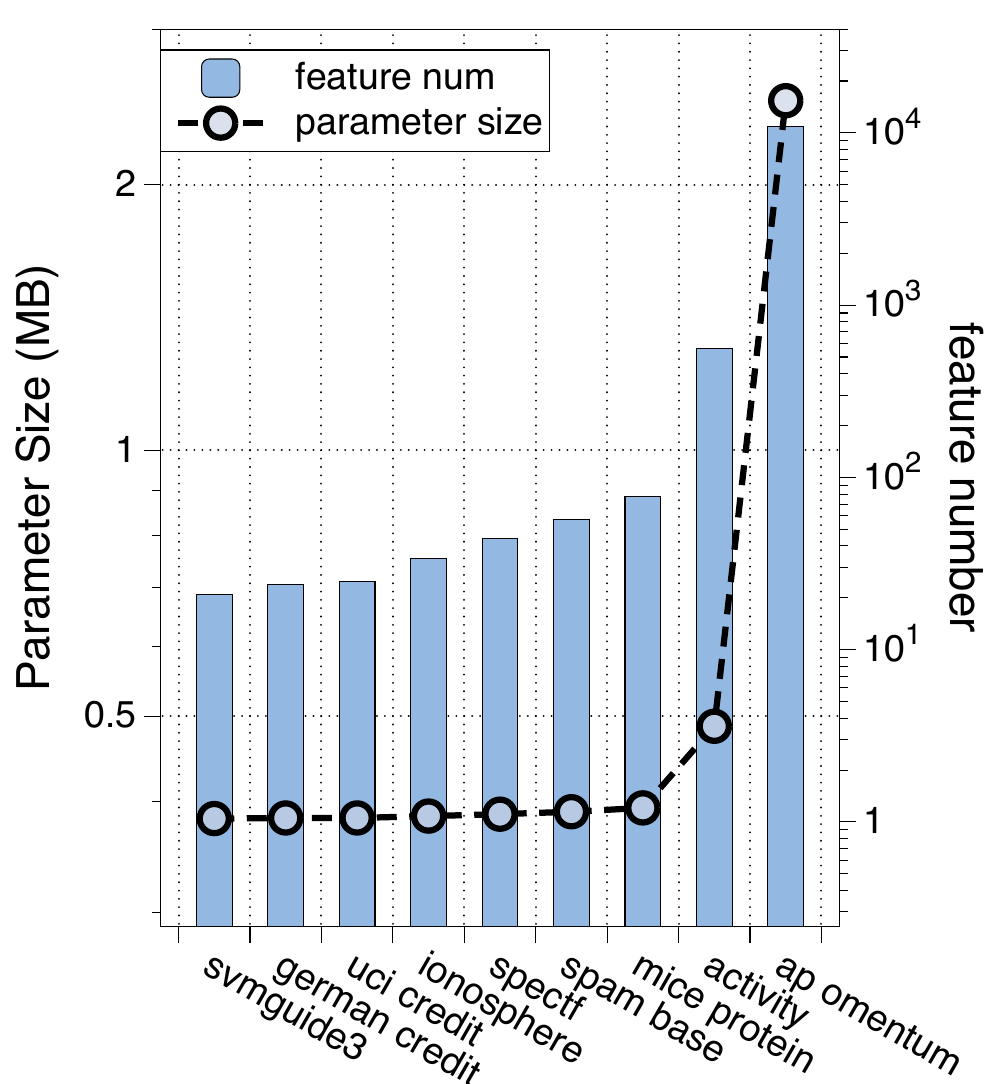}}}
         \subfigure[Data Collection Time]{\includegraphics[width=0.237\textwidth]{{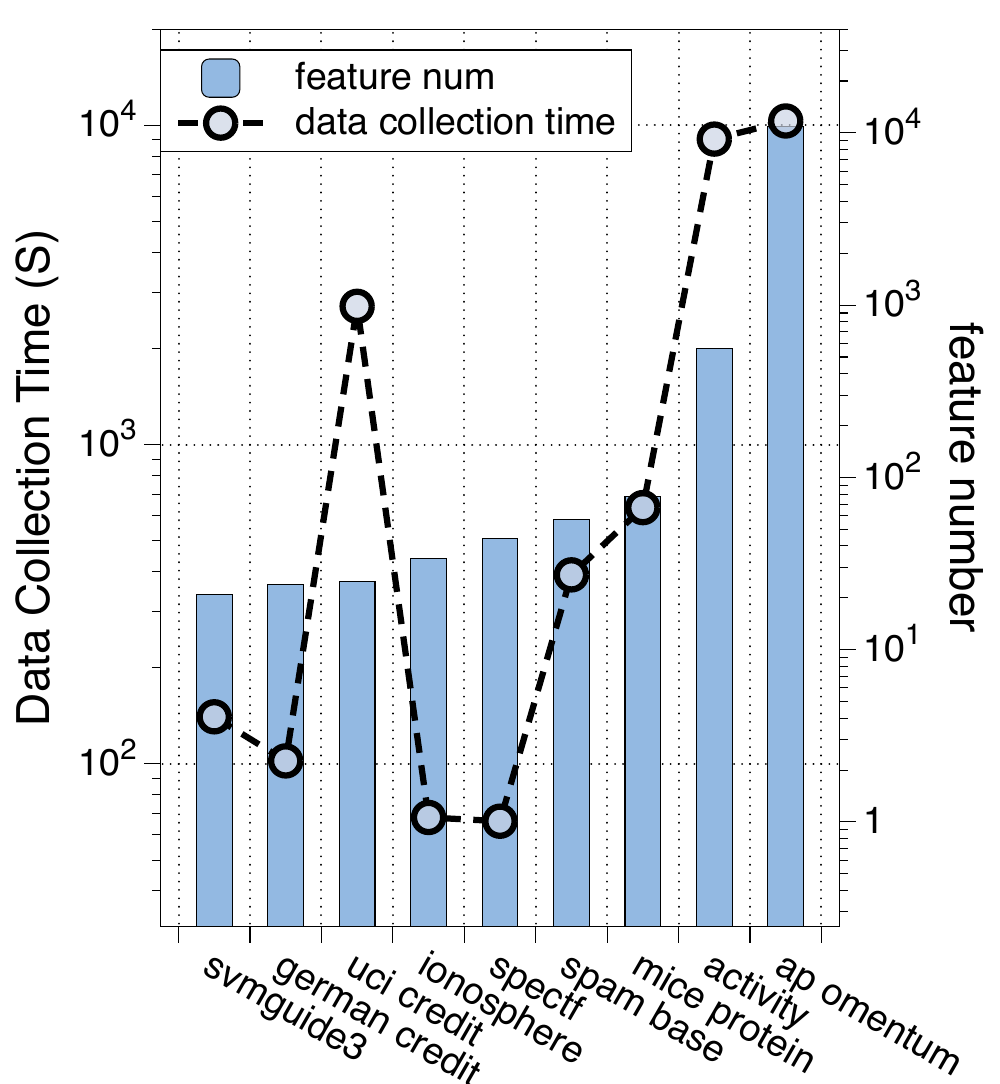}}}
        \caption{Time and Space complexity analysis on classification task in terms of the feature size, training time, inference time, parameter size, and data collection time.}
	\label{exp:C-time-space}
 \vspace{-0.3cm}
\end{figure}
\begin{figure}
	\centering
	\subfigure[Training Time]{\includegraphics[width=0.237\textwidth]{{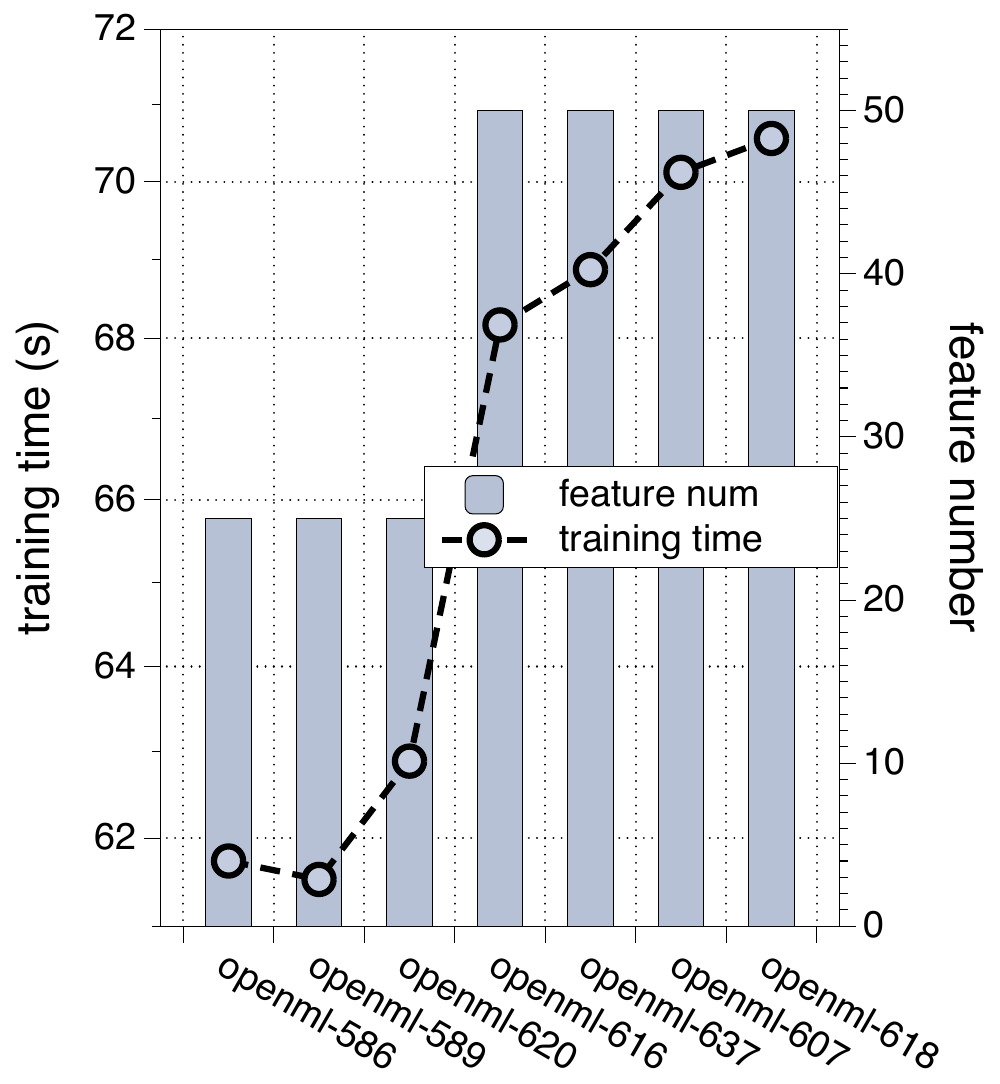}}}
	\subfigure[Inference Time]{\includegraphics[width=0.237\textwidth]{{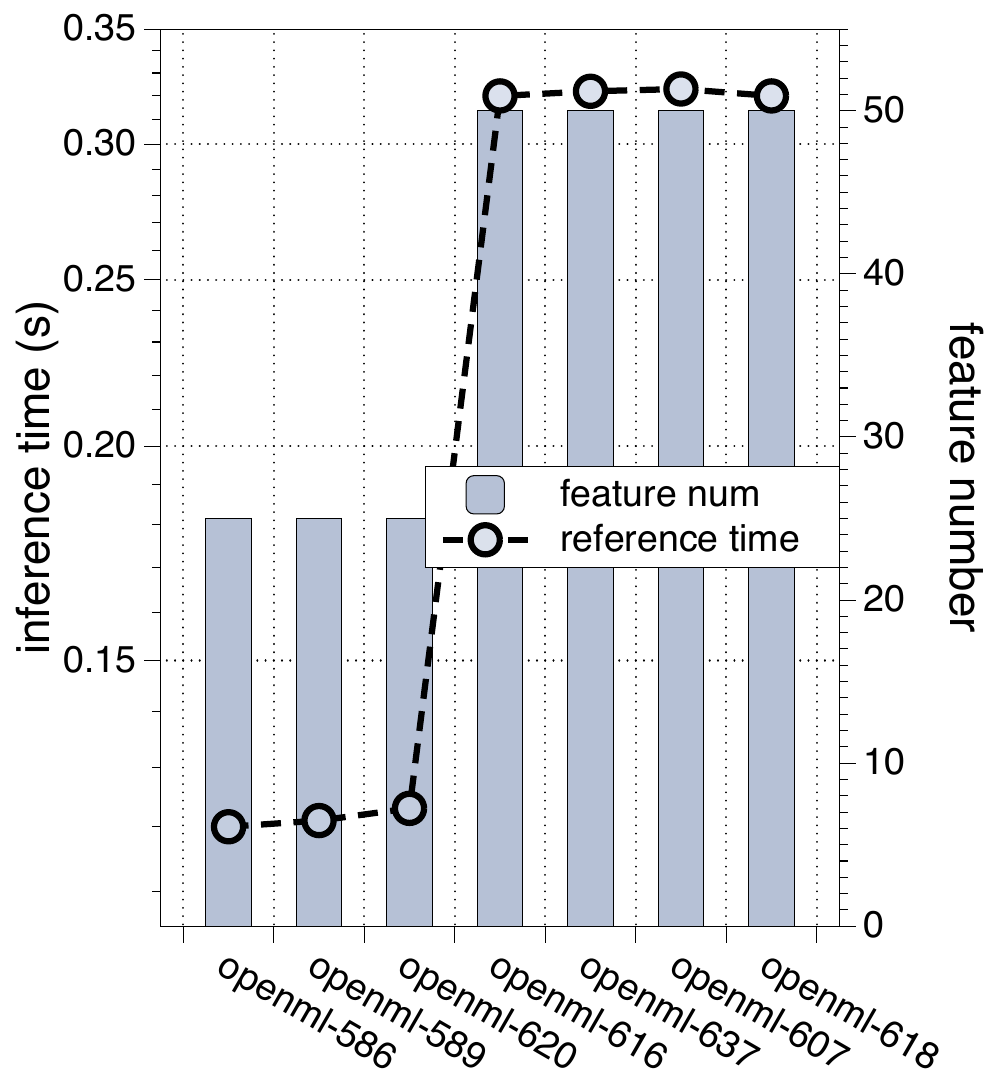}}}
         \subfigure[Parameter Size]{\includegraphics[width=0.237\textwidth]{{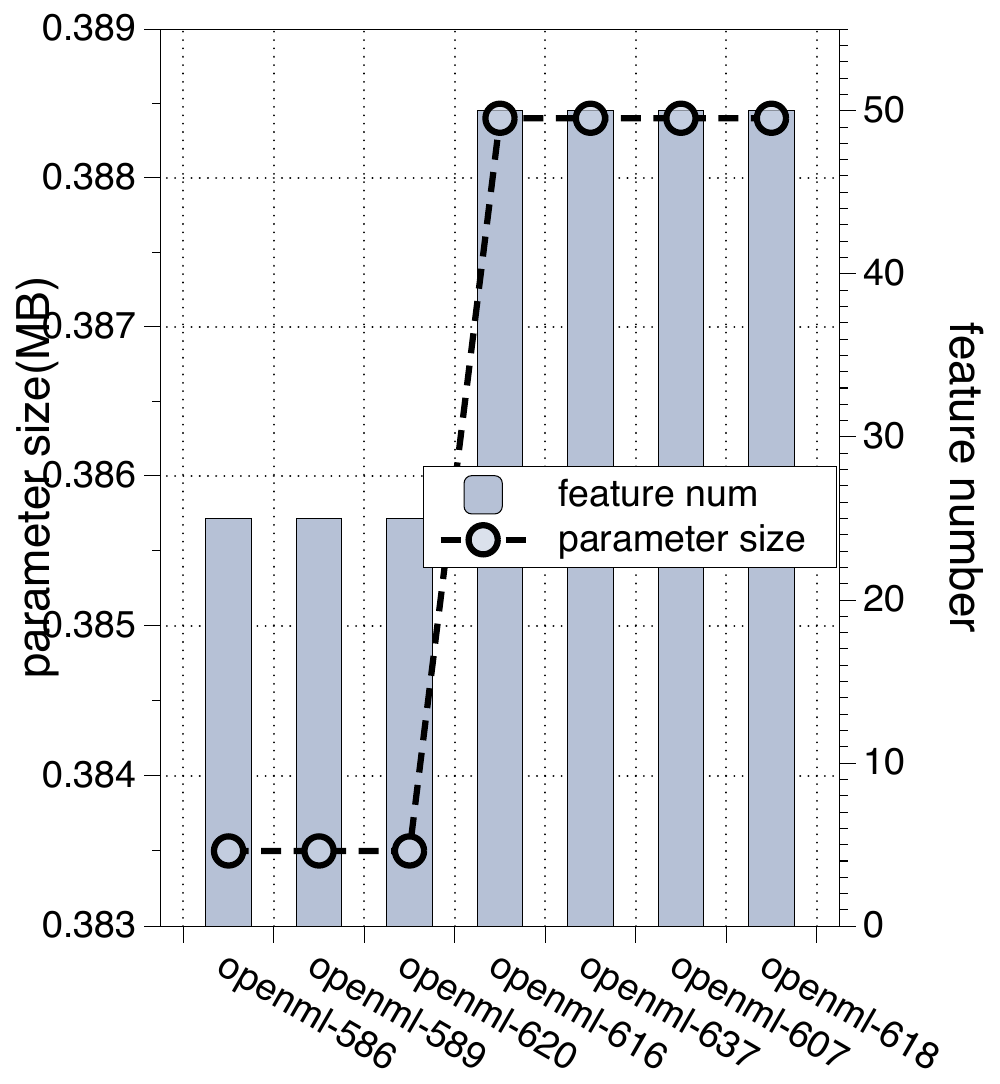}}}
         \subfigure[Data Collection Time]{\includegraphics[width=0.237\textwidth]{{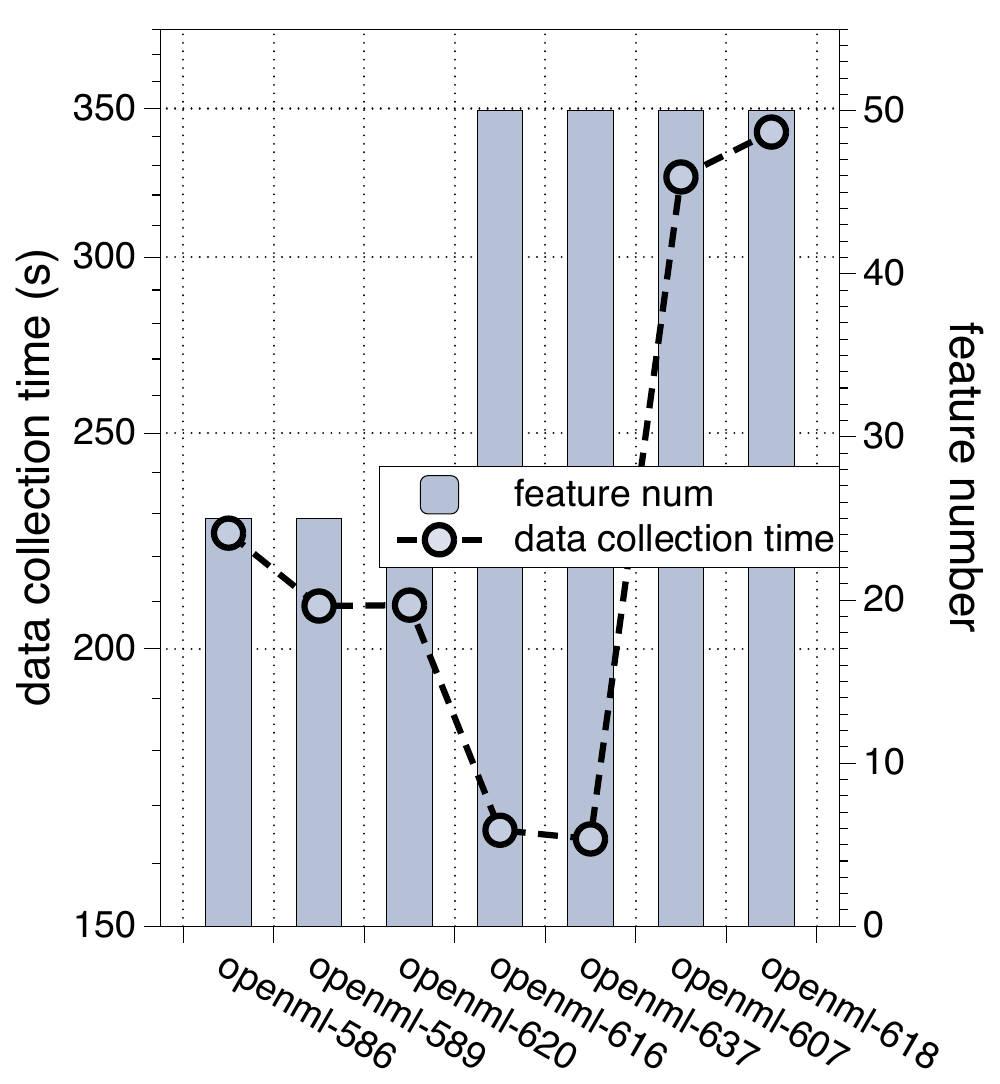}}}
        \caption{Time and Space complexity analysis on regression task in terms of the feature size, training time, inference time, parameter size, and data collection time.}
	\label{exp:R-time-space}
 \vspace{-0.3cm}
\end{figure}

\subsection{Study of the impact of data augmentation.}
Since the order of the feature token sequence does not influence the downstream predictive accuracy, we propose a data augmentation strategy by randomly shuffling the feature token sequence to generate more legal training samples. To assess the impact of data augmentation,  we incrementally increase the number of shufflings and observe its impact on performance improvements.
From Figure \ref{exp:data_augmentation}, we can observe that with the increase of the shuffling number, the downstream ML performance has also been improved across different datasets with great gaps.
A potential explanation for this observation is that the augmentation of shuffling epochs enhances data diversity and volume. These enhancements significantly improve the construction of a distinguishable and informative embedding space, yielding superior feature selection performance.
In summary, the experiment reflects the necessity of the data augmentation strategy in \model\ for keeping good performance.

\subsection{Study of the time and space complexity of \model.}
To assess the time and space complexity of \model, we report \model's training time, inference time, parameter size, and data collection time across all datasets. \textbf{Table \ref{exp:complexity_check}} shows the comparison results. For a more clear comparison, we organized the dataset for comparison based on the feature number and dataset category, as shown in the \textbf{Figure~\ref{exp:C-time-space}} and \textbf{Figure~\ref{exp:R-time-space}}.
In the model training stage,
the model training time and parameter size increase with the growth of the feature number.
We can observe that as the feature number increases from 21 (SVMGuide3) to 10936 (AP\_omentum) (520-fold increase), there is only a 20-fold increase (55.15s to 1118.52s) in the training time and a 7-fold increase (0.3827MB to 2.4894MB) in model size.
In other words, despite the substantial increase in the number of features, the corresponding growth in training time and space complexity is relatively modest.
In the inference stage (from inputting a feature token sequence to outputting the best feature token sequence),
we can observe that the time cost still increases with the growth of the feature number. However, the prediction time in this stage is in the millisecond range, resulting in a very short time despite a huge number of features. The underlying driver is that we embed the feature token sequence into a fixed and low-dimensional embedding, making the gradient-steered optimization process complete within a very short time. 
Thus, this observation indicates that \model\ exhibits exceptional scalability, especially when dealing with high-dimensional feature spaces.
In the data collection stage,
we observe that the time required for reinforcement learning-based data collection increases with the growth of the feature number and sample number. For example, the feature number of the UCI Credit data set is relatively small (25), but the sample number is huge (30,000), resulting in a high data collecting time compared to the dataset of a similar feature number (e.g., the German Credit dataset). The reason is that the RL-based data collector uses a supervised downstream to evaluate the utility of the feature subset in each iteration. The dataset with more samples needs more time to train the downstream ML task. 
Despite taking relatively longer compared to model training, this process is entirely automated, reducing the need for manual intervention. It can learn and adapt to different data collection scenarios, thereby enhancing the adaptability and effectiveness of data collection.

\subsection{Robustness Check.}
To evaluate the robustness of different feature selection algorithms with varying downstream ML models, we replace the random forest model with support vector machine (SVM), XGBoost (XGB), K-nearest neighborhood (KNN), and decision tree (DT). The performance of these algorithms was then evaluated using the SVMGuide3 dataset.
Table \ref{exp:robustness_check} shows the comparison results in terms of F1-score.
We can find that \model\ consistently beats other feature selection baselines regardless of the downstream ML model.
The underlying driver is that \model\ can tailor the feature selection strategy based on the specific characteristics of downstream ML models.
This is achieved by collecting suitable sequential training data that is most suitable for each model type.
Moreover, \model\ embeds feature learning knowledge into a continuous embedding space which enhances its robustness and generalization capability across different ML models. 
In summary, this experiment demonstrates that \model\ can maintain its excellent and stable feature selection performance across different ML models.

\setlength{\tabcolsep}{5mm}{
\begin{table}[t]
\centering
\small
\caption{Analysis of the robustness of different feature selection algorithms using the SVMGuide3 dataset in terms of F1-score.}
% \vspace{-0.2cm}
\begin{tabular}{@{}cccccc@{}}
\toprule \toprule
            & DT            & KNN           & SVM           & XGB           & RF            \\ \midrule
Orininal    & 75.7          & 79.5          & 78.8          & 79.4          & 77.8          \\ \midrule
K-best      & 73.2          & 78.9          & 75.5          & 75.4          & 76.8          \\
mRMR        & 72.0          & 78.5          & 76.3          & 73.4          & 76.8          \\
DNP         & 76.8          & 74.1          & 77.5          & 76.7          & 77.1          \\
DeepPink    & 77.9          & 78.9          & 76.5          & 77.1          & 76.6          \\
KnockoffGAN & 75.5          & 76.7          & 76.9          & 78.6          & 77.9          \\
MCDM        & 73.0          & 78.8          & 76.7          & 75.6          & 76.7          \\
RFE         & 75.5          & 76.7          & 77.5          & 78.8          & 78.1          \\
LASSO       & 70.0          & 74.1          & 77.0          & 78.2          & 77.9          \\
LASSONet    & 71.3          & 73.0          & 76.9          & 75.3          & 76.4          \\
GFS         & 76.8          & 78.9          & 75.2          & 77.1          & 83.1          \\
MARLFS      & 77.9          & 79.9          & 75.4          & 78.6          & 76.8          \\
SARLFS      & 76.0          & 79.2          & 75.3          & 78.9          & 76.2          \\ \midrule
\model         & \textbf{79.0} & \textbf{81.1} & \textbf{78.8} & \textbf{83.8} & \textbf{85.0} \\ \bottomrule \bottomrule
\end{tabular}
\label{exp:robustness_check}
\end{table}}

\begin{figure}[t]
	\centering
	\subfigure[openml\_607]{\label{exp:case:607}\includegraphics[width=0.6\textwidth]{{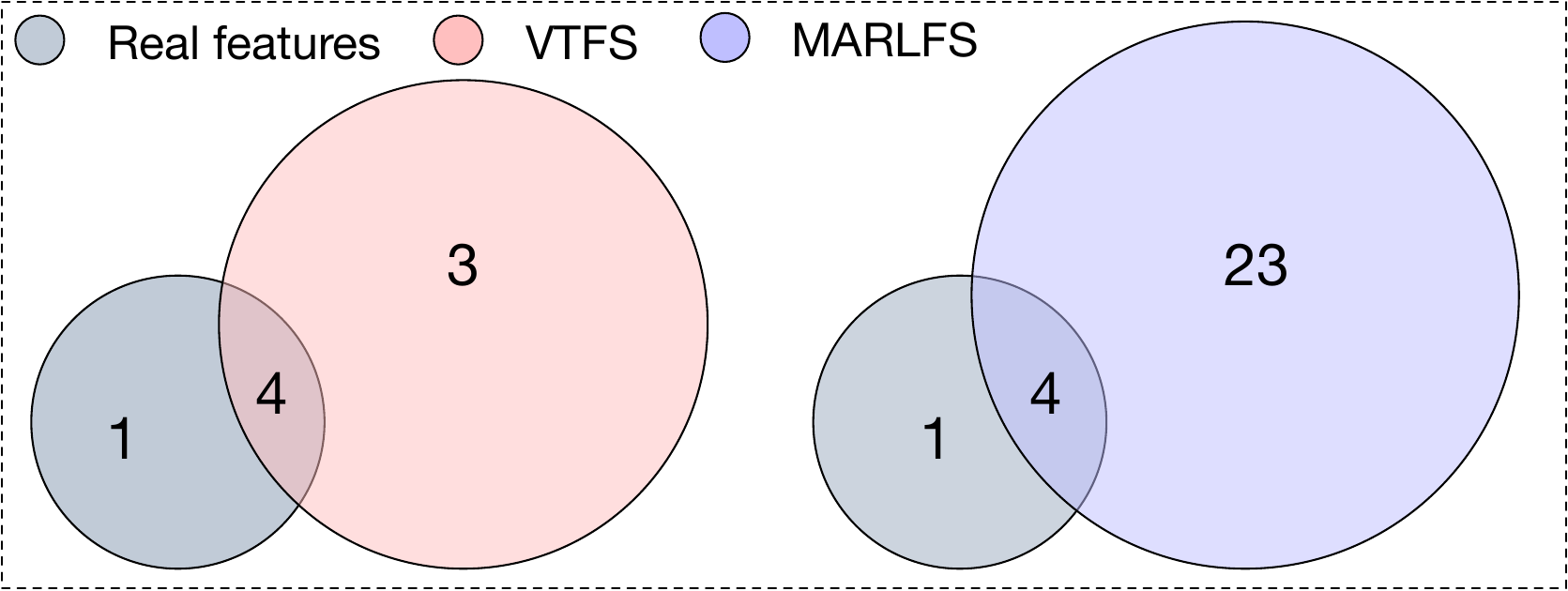}}}
	\subfigure[openml\_618]{\label{exp:case:618}\includegraphics[width=0.6\textwidth]{{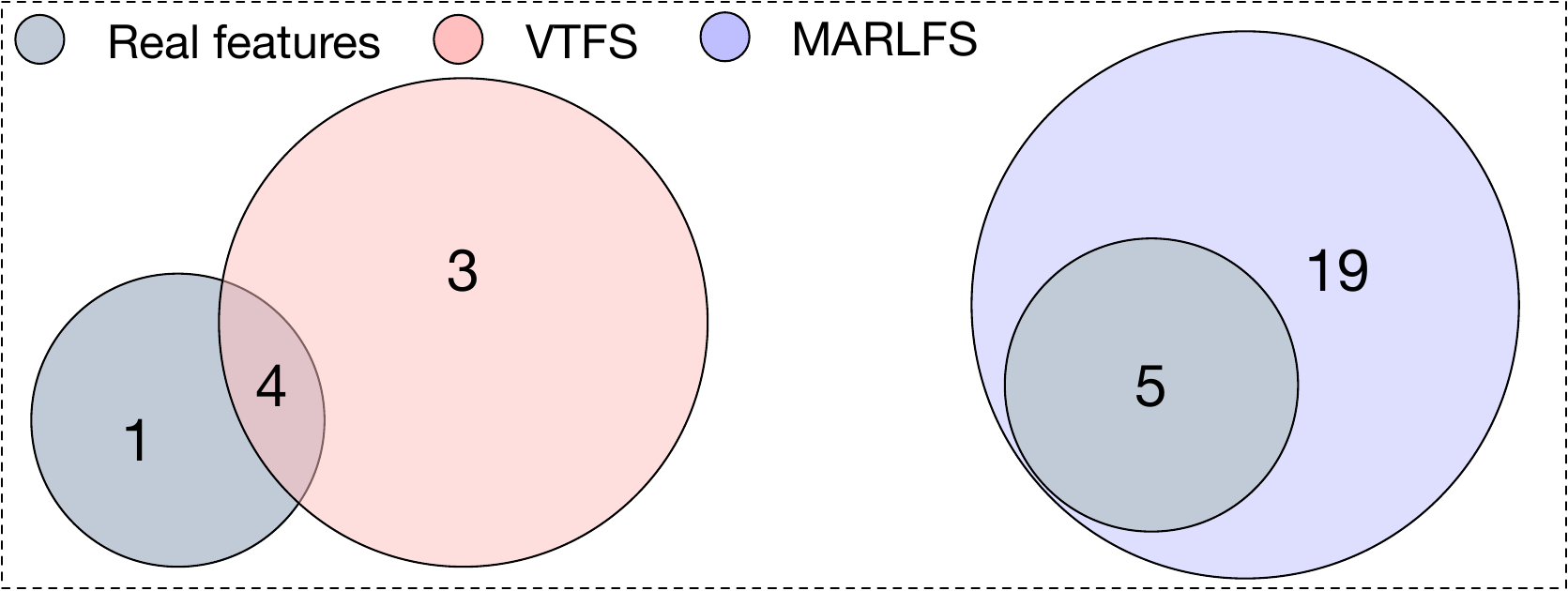}}}
 \vspace{-0.2cm}
	\caption{Case Study: Each dataset consists of 5 real features and 45 fake features. When compared to MARLFS, \model\ demonstrates a superior ability to select feature subsets that are more closely to the real features in both datasets and effectively avoid identifying fake features as real ones.}
	\label{exp:case_study}
\end{figure}

\subsection{Case study: \model\ exhibits noise resistance and quality feature attention.} 
The OpenML datasets are simulated by human experts. So we know the real relevant features within these datasets.
Thus, we design a case study to show the overlap between the selected features and the real ones.
Here, we take openml\_607 and openml\_618 datasets as examples.
Both of them have 5 real features and 45 fake features.
We employ MARLFS~\cite{marlfs} to serve as a comparative model alongside \model.
Figure~\ref{exp:case_study} shows the comparison results.
Regarding the openml\_607 dataset, we can find that  \model\ selects 7 features, of which 4 are real and 3 are fake. In contrast, MARLFS selects 27 features, with only 4 being real and the remaining 23 being fake.
For the openml\_618 dataset, \model\ maintains a similar performance. While MARLFS successfully identifies all real features, it also includes 19 fake features in its selection.
These observations indicate that, in comparison to MARLFS, \model\ is more effective at understanding the complex relationships within the feature space. As a result, it is able to produce a feature subset that more closely aligns with the actual features, thereby reducing the likelihood of making false-positive errors.
In summary, this case study demonstrates that \model\ exhibits robustness in filtering out noise within the feature space and is capable of producing high-quality and reliable feature subsets.

\section{Related Work}

Feature selection methods can be divided into three categories according to the selection strategies~\cite{li2017feature}: 1) filter methods; 2) wrapper methods; 3) embedded methods. 

The filter methods~\cite{kbest,mrmr,he2005laplacian} evaluate features by calculating the correlation between features based on statistical properties of data, and selects the feature subset with the highest score. 
Univariate statistical tests, such as variance analysis F-test~\cite{f-test}, are widely used in filter methods. The F-statistic values are used as ranking scores for each feature, where higher F-statistic values correspond to more important features. Other classical statistical methods, including Student's t-test~\cite{t-test}, Pearson correlation test~\cite{pearson}, chi-square test~\cite{chi-square}, Kolmogorov-Smirnov test~\cite{Kolmogorov-Smirnov}, Wilks lambda test~\cite{wilklabmda}, and Wilcoxon signed-rank test~\cite{wilcoxon}, can be similarly applied to feature selection. These methods have low computational complexity and can efficiently select feature subsets from high-dimensional datasets. However, they ignore the dependency and interaction among features, potentially leading to suboptimal results.

The wrapper methods~\cite{gfe,marlfs,sarlfs,liu2023interactive} are based on a specific dataset, define a machine learning model in advance, and iteratively evaluate the candidate feature subset. For instance, reinforcement learning-based methods model the feature selection process with a multi-agent system, where agents decide whether to select a particular feature, optimize the utility of selected feature subsets, and use the utility and feature redundancy as reward feedback in each iteration. These methods often outperform filter methods as they enumerate various combinations of feature subsets. However, due to the need to enumerate all possible feature subsets, it is an NP-hard problem, and the evaluation using downstream machine learning models after each iteration leads to lower computational efficiency. These methods may suffer from convergence difficulties and instability, potentially making it difficult to identify the optimal feature subset.

The embedded methods~\cite{lasso,lassonet,rfe,C2IMUFS} transform the feature selection task into a regularization term in the prediction loss of a machine learning model to accelerate the selection process. For example, LASSO assumes a linear dependency between input features and output, penalizing the L1 norm of feature weights. Lasso produces a sparse solution where the weights of irrelevant features are set to zero. However, Lasso fails to capture nonlinear dependencies.
The three types of methods have excellent performance on specific machine learning models. However, the filter and embedded methods exhibit limited generalization ability over various domain datasets and downstream predictive models.  
The wrapper methods suffer from large search space and cannot ensure the identification of global optimal. 

In addition, other studies have proposed two types of hybrid feature selection methods: 1) homogeneous methods~\cite{seijo2017testing,pes2017exploiting,seijo2017ensemble}; 2) heterogeneous methods~\cite{haque2016heterogeneous,seijo2019developing}. However, these methods are limited by the basic aggregation strategies.
Thus, it is critical to develop a new research perspective to enhance the generalization and effectiveness. 
In contrast to the above existing works, we propose a novel generative AI perspective that embeds the knowledge of feature selection into a continuous embedding space, then effectively identifies feature subsets using the gradient-steered search and autoregressive generation. 
\section{Conclusion}

This paper explores a new research perspective on the feature selection problem: embedding feature selection knowledge into a continuous space and generating the best feature subsets based on a gradient-ascent search method. We implement a three-step framework to map feature subset into an embedding space for optimizing feature selection: 
1) We develop a deep variational transformer-based encoder-decoder-evaluator framework to learn a continuous embedding space that can map feature subsets into embedding vectors associated with utility scores. 
2) We leverage the well-trained feature subset utility evaluator as a gradient provider to identify the optimal feature subset embedding.
3) We decode the optimal feature subset embedding to generate the best feature subset in an autoregressive manner. 
Our research findings indicate that: 1) the encoder-decoder-evaluator framework effectively constructs the feature subset embedding space and maintains the utility of feature subsets; 2) the gradient-based search strategy generates gradient and direction information to effectively steer the gradient ascent-based search and identify the optimal feature subset.
In the future, we aim to enhance the generalization capability of \model\ across various domains, scenarios, and distributions.

\bibliographystyle{ACM-Reference-Format}
\bibliography{tkdd}


\begin{thebibliography}{42}

%%% the Web address.

\ifx \showCODEN    \undefined \def \showCODEN     #1{\unskip}     \fi
\ifx \showDOI      \undefined \def \showDOI       #1{#1}\fi
\ifx \showISBNx    \undefined \def \showISBNx     #1{\unskip}     \fi
\ifx \showISBNxiii \undefined \def \showISBNxiii  #1{\unskip}     \fi
\ifx \showISSN     \undefined \def \showISSN      #1{\unskip}     \fi
\ifx \showLCCN     \undefined \def \showLCCN      #1{\unskip}     \fi
\ifx \shownote     \undefined \def \shownote      #1{#1}          \fi
\ifx \showarticletitle \undefined \def \showarticletitle #1{#1}   \fi
\ifx \showURL      \undefined \def \showURL       {\relax}        \fi
% invisible to TeX
\providecommand\bibfield[2]{#2}
\providecommand\bibinfo[2]{#2}
\providecommand\natexlab[1]{#1}
\providecommand\showeprint[2][]{arXiv:#2}

\bibitem[Cand{\`e}s et~al\mbox{.}(2016)]%
        {knockoff}
\bibfield{author}{\bibinfo{person}{Emmanuel~Jean Cand{\`e}s}, \bibinfo{person}{Yingying Fan}, \bibinfo{person}{Lucas Janson}, {and} \bibinfo{person}{Jinchi Lv}.} \bibinfo{year}{2016}\natexlab{}.
\newblock \bibinfo{booktitle}{\emph{Panning for gold: Model-free knockoffs for high-dimensional controlled variable selection}}. Vol.~\bibinfo{volume}{1610}.
\newblock \bibinfo{publisher}{Department of Statistics, Stanford University Stanford, CA, USA}.
\newblock


\bibitem[Elssied et~al\mbox{.}(2014)]%
        {f-test}
\bibfield{author}{\bibinfo{person}{Nadir Omer~Fadl Elssied}, \bibinfo{person}{Othman Ibrahim}, {and} \bibinfo{person}{Ahmed~Hamza Osman}.} \bibinfo{year}{2014}\natexlab{}.
\newblock \showarticletitle{A novel feature selection based on one-way anova f-test for e-mail spam classification}.
\newblock \bibinfo{journal}{\emph{Research Journal of Applied Sciences, Engineering and Technology}} \bibinfo{volume}{7}, \bibinfo{number}{3} (\bibinfo{year}{2014}), \bibinfo{pages}{625--638}.
\newblock


\bibitem[Fan et~al\mbox{.}(2021)]%
        {fan2021autogfs}
\bibfield{author}{\bibinfo{person}{Wei Fan}, \bibinfo{person}{Kunpeng Liu}, \bibinfo{person}{Hao Liu}, \bibinfo{person}{Ahmad Hariri}, \bibinfo{person}{Dejing Dou}, {and} \bibinfo{person}{Yanjie Fu}.} \bibinfo{year}{2021}\natexlab{}.
\newblock \showarticletitle{Autogfs: Automated group-based feature selection via interactive reinforcement learning}. In \bibinfo{booktitle}{\emph{Proceedings of the 2021 SIAM International Conference on Data Mining (SDM)}}. SIAM, \bibinfo{pages}{342--350}.
\newblock


\bibitem[Forman et~al\mbox{.}(2003)]%
        {forman2003extensive}
\bibfield{author}{\bibinfo{person}{George Forman} {et~al\mbox{.}}} \bibinfo{year}{2003}\natexlab{}.
\newblock \showarticletitle{An extensive empirical study of feature selection metrics for text classification.}
\newblock \bibinfo{journal}{\emph{J. Mach. Learn. Res.}} \bibinfo{volume}{3}, \bibinfo{number}{Mar} (\bibinfo{year}{2003}), \bibinfo{pages}{1289--1305}.
\newblock


\bibitem[Granitto et~al\mbox{.}(2006)]%
        {rfe}
\bibfield{author}{\bibinfo{person}{Pablo~M Granitto}, \bibinfo{person}{Cesare Furlanello}, \bibinfo{person}{Franco Biasioli}, {and} \bibinfo{person}{Flavia Gasperi}.} \bibinfo{year}{2006}\natexlab{}.
\newblock \showarticletitle{Recursive feature elimination with random forest for PTR-MS analysis of agroindustrial products}.
\newblock \bibinfo{journal}{\emph{Chemometrics and intelligent laboratory systems}} \bibinfo{volume}{83}, \bibinfo{number}{2} (\bibinfo{year}{2006}), \bibinfo{pages}{83--90}.
\newblock


\bibitem[Hall(1999)]%
        {hall1999feature}
\bibfield{author}{\bibinfo{person}{Mark~A Hall}.} \bibinfo{year}{1999}\natexlab{}.
\newblock \showarticletitle{Feature selection for discrete and numeric class machine learning}.
\newblock  (\bibinfo{year}{1999}).
\newblock


\bibitem[Haque et~al\mbox{.}(2016)]%
        {haque2016heterogeneous}
\bibfield{author}{\bibinfo{person}{Mohammad~Nazmul Haque}, \bibinfo{person}{Nasimul Noman}, \bibinfo{person}{Regina Berretta}, {and} \bibinfo{person}{Pablo Moscato}.} \bibinfo{year}{2016}\natexlab{}.
\newblock \showarticletitle{Heterogeneous ensemble combination search using genetic algorithm for class imbalanced data classification}.
\newblock \bibinfo{journal}{\emph{PloS one}} \bibinfo{volume}{11}, \bibinfo{number}{1} (\bibinfo{year}{2016}), \bibinfo{pages}{e0146116}.
\newblock


\bibitem[Hashemi et~al\mbox{.}(2022)]%
        {mcdm}
\bibfield{author}{\bibinfo{person}{Amin Hashemi}, \bibinfo{person}{Mohammad~Bagher Dowlatshahi}, {and} \bibinfo{person}{Hossein Nezamabadi-pour}.} \bibinfo{year}{2022}\natexlab{}.
\newblock \showarticletitle{Ensemble of feature selection algorithms: a multi-criteria decision-making approach}.
\newblock \bibinfo{journal}{\emph{International Journal of Machine Learning and Cybernetics}} \bibinfo{volume}{13}, \bibinfo{number}{1} (\bibinfo{year}{2022}), \bibinfo{pages}{49--69}.
\newblock


\bibitem[He et~al\mbox{.}(2005)]%
        {he2005laplacian}
\bibfield{author}{\bibinfo{person}{Xiaofei He}, \bibinfo{person}{Deng Cai}, {and} \bibinfo{person}{Partha Niyogi}.} \bibinfo{year}{2005}\natexlab{}.
\newblock \showarticletitle{Laplacian score for feature selection}.
\newblock \bibinfo{journal}{\emph{Advances in neural information processing systems}}  \bibinfo{volume}{18} (\bibinfo{year}{2005}).
\newblock


\bibitem[Hochreiter and Schmidhuber(1997)]%
        {LSTM}
\bibfield{author}{\bibinfo{person}{Sepp Hochreiter} {and} \bibinfo{person}{J{\"u}rgen Schmidhuber}.} \bibinfo{year}{1997}\natexlab{}.
\newblock \showarticletitle{Long short-term memory}.
\newblock \bibinfo{journal}{\emph{Neural computation}} \bibinfo{volume}{9}, \bibinfo{number}{8} (\bibinfo{year}{1997}), \bibinfo{pages}{1735--1780}.
\newblock


\bibitem[Huang et~al\mbox{.}(2023)]%
        {C2IMUFS}
\bibfield{author}{\bibinfo{person}{Yanyong Huang}, \bibinfo{person}{Zongxin Shen}, \bibinfo{person}{Yuxin Cai}, \bibinfo{person}{Xiuwen Yi}, \bibinfo{person}{Dongjie Wang}, \bibinfo{person}{Fengmao Lv}, {and} \bibinfo{person}{Tianrui Li}.} \bibinfo{year}{2023}\natexlab{}.
\newblock \showarticletitle{C2IMUFS: Complementary and Consensus Learning-Based Incomplete Multi-View Unsupervised Feature Selection}.
\newblock \bibinfo{journal}{\emph{IEEE Transactions on Knowledge and Data Engineering}} \bibinfo{volume}{35}, \bibinfo{number}{10} (\bibinfo{year}{2023}), \bibinfo{pages}{10681--10694}.
\newblock
\urldef\tempurl%
\url{https://doi.org/10.1109/TKDE.2023.3266595}
\showDOI{\tempurl}


\bibitem[Hupse and Karssemeijer(2010)]%
        {wilklabmda}
\bibfield{author}{\bibinfo{person}{Rianne Hupse} {and} \bibinfo{person}{Nico Karssemeijer}.} \bibinfo{year}{2010}\natexlab{}.
\newblock \showarticletitle{The effect of feature selection methods on computer-aided detection of masses in mammograms}.
\newblock \bibinfo{journal}{\emph{Physics in Medicine \& Biology}} \bibinfo{volume}{55}, \bibinfo{number}{10} (\bibinfo{year}{2010}), \bibinfo{pages}{2893}.
\newblock


\bibitem[Ivanov and Riccardi(2012)]%
        {Kolmogorov-Smirnov}
\bibfield{author}{\bibinfo{person}{Alexei Ivanov} {and} \bibinfo{person}{Giuseppe Riccardi}.} \bibinfo{year}{2012}\natexlab{}.
\newblock \showarticletitle{Kolmogorov-Smirnov test for feature selection in emotion recognition from speech}. In \bibinfo{booktitle}{\emph{2012 IEEE international conference on acoustics, speech and signal processing (ICASSP)}}. IEEE, \bibinfo{pages}{5125--5128}.
\newblock


\bibitem[Jordon et~al\mbox{.}(2018)]%
        {knockoffgan}
\bibfield{author}{\bibinfo{person}{James Jordon}, \bibinfo{person}{Jinsung Yoon}, {and} \bibinfo{person}{Mihaela van~der Schaar}.} \bibinfo{year}{2018}\natexlab{}.
\newblock \showarticletitle{KnockoffGAN: Generating knockoffs for feature selection using generative adversarial networks}. In \bibinfo{booktitle}{\emph{International conference on learning representations}}.
\newblock


\bibitem[Kim et~al\mbox{.}(2000)]%
        {kim2000feature}
\bibfield{author}{\bibinfo{person}{YeongSeog Kim}, \bibinfo{person}{W~Nick Street}, {and} \bibinfo{person}{Filippo Menczer}.} \bibinfo{year}{2000}\natexlab{}.
\newblock \showarticletitle{Feature selection in unsupervised learning via evolutionary search}. In \bibinfo{booktitle}{\emph{Proceedings of the sixth ACM SIGKDD international conference on Knowledge discovery and data mining}}. \bibinfo{pages}{365--369}.
\newblock


\bibitem[Kingma and Welling(2013)]%
        {vae}
\bibfield{author}{\bibinfo{person}{Diederik~P Kingma} {and} \bibinfo{person}{Max Welling}.} \bibinfo{year}{2013}\natexlab{}.
\newblock \showarticletitle{Auto-encoding variational bayes}.
\newblock \bibinfo{journal}{\emph{arXiv preprint arXiv:1312.6114}} (\bibinfo{year}{2013}).
\newblock


\bibitem[Kohavi and John(1997)]%
        {kohavi1997wrappers}
\bibfield{author}{\bibinfo{person}{Ron Kohavi} {and} \bibinfo{person}{George~H John}.} \bibinfo{year}{1997}\natexlab{}.
\newblock \showarticletitle{Wrappers for feature subset selection}.
\newblock \bibinfo{journal}{\emph{Artificial intelligence}} \bibinfo{volume}{97}, \bibinfo{number}{1-2} (\bibinfo{year}{1997}), \bibinfo{pages}{273--324}.
\newblock


\bibitem[Leardi(1996)]%
        {gfe}
\bibfield{author}{\bibinfo{person}{Riccardo Leardi}.} \bibinfo{year}{1996}\natexlab{}.
\newblock \showarticletitle{Genetic algorithms in feature selection}.
\newblock In \bibinfo{booktitle}{\emph{Genetic algorithms in molecular modeling}}. \bibinfo{publisher}{Elsevier}, \bibinfo{pages}{67--86}.
\newblock


\bibitem[Lemhadri et~al\mbox{.}(2021)]%
        {lassonet}
\bibfield{author}{\bibinfo{person}{Ismael Lemhadri}, \bibinfo{person}{Feng Ruan}, {and} \bibinfo{person}{Rob Tibshirani}.} \bibinfo{year}{2021}\natexlab{}.
\newblock \showarticletitle{Lassonet: Neural networks with feature sparsity}. In \bibinfo{booktitle}{\emph{International Conference on Artificial Intelligence and Statistics}}. PMLR, \bibinfo{pages}{10--18}.
\newblock


\bibitem[Li et~al\mbox{.}(2017)]%
        {li2017feature}
\bibfield{author}{\bibinfo{person}{Jundong Li}, \bibinfo{person}{Kewei Cheng}, \bibinfo{person}{Suhang Wang}, \bibinfo{person}{Fred Morstatter}, \bibinfo{person}{Robert~P Trevino}, \bibinfo{person}{Jiliang Tang}, {and} \bibinfo{person}{Huan Liu}.} \bibinfo{year}{2017}\natexlab{}.
\newblock \showarticletitle{Feature selection: A data perspective}.
\newblock \bibinfo{journal}{\emph{ACM Computing Surveys (CSUR)}} \bibinfo{volume}{50}, \bibinfo{number}{6} (\bibinfo{year}{2017}), \bibinfo{pages}{1--45}.
\newblock


\bibitem[Liu et~al\mbox{.}(2017)]%
        {DNP}
\bibfield{author}{\bibinfo{person}{Bo Liu}, \bibinfo{person}{Ying Wei}, \bibinfo{person}{Yu Zhang}, {and} \bibinfo{person}{Qiang Yang}.} \bibinfo{year}{2017}\natexlab{}.
\newblock \showarticletitle{Deep Neural Networks for High Dimension, Low Sample Size Data.}. In \bibinfo{booktitle}{\emph{IJCAI}}. \bibinfo{pages}{2287--2293}.
\newblock


\bibitem[Liu et~al\mbox{.}(2019)]%
        {marlfs}
\bibfield{author}{\bibinfo{person}{Kunpeng Liu}, \bibinfo{person}{Yanjie Fu}, \bibinfo{person}{Pengfei Wang}, \bibinfo{person}{Le Wu}, \bibinfo{person}{Rui Bo}, {and} \bibinfo{person}{Xiaolin Li}.} \bibinfo{year}{2019}\natexlab{}.
\newblock \showarticletitle{Automating feature subspace exploration via multi-agent reinforcement learning}. In \bibinfo{booktitle}{\emph{Proceedings of the 25th ACM SIGKDD International Conference on Knowledge Discovery \& Data Mining}}. \bibinfo{pages}{207--215}.
\newblock


\bibitem[Liu et~al\mbox{.}(2023)]%
        {liu2023interactive}
\bibfield{author}{\bibinfo{person}{Kunpeng Liu}, \bibinfo{person}{Dongjie Wang}, \bibinfo{person}{Wan Du}, \bibinfo{person}{Dapeng~Oliver Wu}, {and} \bibinfo{person}{Yanjie Fu}.} \bibinfo{year}{2023}\natexlab{}.
\newblock \showarticletitle{Interactive reinforced feature selection with traverse strategy}.
\newblock \bibinfo{journal}{\emph{Knowledge and Information Systems}} \bibinfo{volume}{65}, \bibinfo{number}{5} (\bibinfo{year}{2023}), \bibinfo{pages}{1935--1962}.
\newblock


\bibitem[Liu et~al\mbox{.}(2021)]%
        {sarlfs}
\bibfield{author}{\bibinfo{person}{Kunpeng Liu}, \bibinfo{person}{Pengfei Wang}, \bibinfo{person}{Dongjie Wang}, \bibinfo{person}{Wan Du}, \bibinfo{person}{Dapeng~Oliver Wu}, {and} \bibinfo{person}{Yanjie Fu}.} \bibinfo{year}{2021}\natexlab{}.
\newblock \showarticletitle{Efficient Reinforced Feature Selection via Early Stopping Traverse Strategy}. In \bibinfo{booktitle}{\emph{2021 IEEE International Conference on Data Mining (ICDM)}}. IEEE, \bibinfo{pages}{399--408}.
\newblock


\bibitem[Liu et~al\mbox{.}(2020)]%
        {pearson}
\bibfield{author}{\bibinfo{person}{Yaqing Liu}, \bibinfo{person}{Yong Mu}, \bibinfo{person}{Keyu Chen}, \bibinfo{person}{Yiming Li}, {and} \bibinfo{person}{Jinghuan Guo}.} \bibinfo{year}{2020}\natexlab{}.
\newblock \showarticletitle{Daily activity feature selection in smart homes based on pearson correlation coefficient}.
\newblock \bibinfo{journal}{\emph{Neural Processing Letters}}  \bibinfo{volume}{51} (\bibinfo{year}{2020}), \bibinfo{pages}{1771--1787}.
\newblock


\bibitem[Lu et~al\mbox{.}(2018)]%
        {deeppink}
\bibfield{author}{\bibinfo{person}{Yang Lu}, \bibinfo{person}{Yingying Fan}, \bibinfo{person}{Jinchi Lv}, {and} \bibinfo{person}{William Stafford~Noble}.} \bibinfo{year}{2018}\natexlab{}.
\newblock \showarticletitle{DeepPINK: reproducible feature selection in deep neural networks}.
\newblock \bibinfo{journal}{\emph{Advances in neural information processing systems}}  \bibinfo{volume}{31} (\bibinfo{year}{2018}).
\newblock


\bibitem[Mnih et~al\mbox{.}(2015)]%
        {DQN}
\bibfield{author}{\bibinfo{person}{Volodymyr Mnih}, \bibinfo{person}{Koray Kavukcuoglu}, \bibinfo{person}{David Silver}, \bibinfo{person}{Andrei~A Rusu}, \bibinfo{person}{Joel Veness}, \bibinfo{person}{Marc~G Bellemare}, \bibinfo{person}{Alex Graves}, \bibinfo{person}{Martin Riedmiller}, \bibinfo{person}{Andreas~K Fidjeland}, \bibinfo{person}{Georg Ostrovski}, {et~al\mbox{.}}} \bibinfo{year}{2015}\natexlab{}.
\newblock \showarticletitle{Human-level control through deep reinforcement learning}.
\newblock \bibinfo{journal}{\emph{nature}} \bibinfo{volume}{518}, \bibinfo{number}{7540} (\bibinfo{year}{2015}), \bibinfo{pages}{529--533}.
\newblock


\bibitem[Narendra and Fukunaga(1977)]%
        {narendra1977branch}
\bibfield{author}{\bibinfo{person}{Patrenahalli~M. Narendra} {and} \bibinfo{person}{Keinosuke Fukunaga}.} \bibinfo{year}{1977}\natexlab{}.
\newblock \showarticletitle{A branch and bound algorithm for feature subset selection}.
\newblock \bibinfo{journal}{\emph{IEEE Transactions on computers}} \bibinfo{number}{9} (\bibinfo{year}{1977}), \bibinfo{pages}{917--922}.
\newblock


\bibitem[Peng et~al\mbox{.}(2005)]%
        {mrmr}
\bibfield{author}{\bibinfo{person}{Hanchuan Peng}, \bibinfo{person}{Fuhui Long}, {and} \bibinfo{person}{Chris Ding}.} \bibinfo{year}{2005}\natexlab{}.
\newblock \showarticletitle{Feature selection based on mutual information criteria of max-dependency, max-relevance, and min-redundancy}.
\newblock \bibinfo{journal}{\emph{IEEE Transactions on pattern analysis and machine intelligence}} \bibinfo{volume}{27}, \bibinfo{number}{8} (\bibinfo{year}{2005}), \bibinfo{pages}{1226--1238}.
\newblock


\bibitem[Pes et~al\mbox{.}(2017)]%
        {pes2017exploiting}
\bibfield{author}{\bibinfo{person}{Barbara Pes}, \bibinfo{person}{Nicoletta Dess{\`\i}}, {and} \bibinfo{person}{Marta Angioni}.} \bibinfo{year}{2017}\natexlab{}.
\newblock \showarticletitle{Exploiting the ensemble paradigm for stable feature selection: a case study on high-dimensional genomic data}.
\newblock \bibinfo{journal}{\emph{Information Fusion}}  \bibinfo{volume}{35} (\bibinfo{year}{2017}), \bibinfo{pages}{132--147}.
\newblock


\bibitem[Sayeedunnisa et~al\mbox{.}(2018)]%
        {wilcoxon}
\bibfield{author}{\bibinfo{person}{S~Fouzia Sayeedunnisa}, \bibinfo{person}{Nagaratna~P Hegde}, {and} \bibinfo{person}{Khaleel Ur~Rahman Khan}.} \bibinfo{year}{2018}\natexlab{}.
\newblock \showarticletitle{Wilcoxon signed rank based feature selection for sentiment classification}. In \bibinfo{booktitle}{\emph{Proceedings of the Second International Conference on Computational Intelligence and Informatics: ICCII 2017}}. Springer, \bibinfo{pages}{293--310}.
\newblock


\bibitem[Seijo-Pardo et~al\mbox{.}(2017a)]%
        {seijo2017testing}
\bibfield{author}{\bibinfo{person}{Borja Seijo-Pardo}, \bibinfo{person}{Ver{\'o}nica Bol{\'o}n-Canedo}, {and} \bibinfo{person}{Amparo Alonso-Betanzos}.} \bibinfo{year}{2017}\natexlab{a}.
\newblock \showarticletitle{Testing different ensemble configurations for feature selection}.
\newblock \bibinfo{journal}{\emph{Neural Processing Letters}} \bibinfo{volume}{46}, \bibinfo{number}{3} (\bibinfo{year}{2017}), \bibinfo{pages}{857--880}.
\newblock


\bibitem[Seijo-Pardo et~al\mbox{.}(2019)]%
        {seijo2019developing}
\bibfield{author}{\bibinfo{person}{Borja Seijo-Pardo}, \bibinfo{person}{Ver{\'o}nica Bol{\'o}n-Canedo}, {and} \bibinfo{person}{Amparo Alonso-Betanzos}.} \bibinfo{year}{2019}\natexlab{}.
\newblock \showarticletitle{On developing an automatic threshold applied to feature selection ensembles}.
\newblock \bibinfo{journal}{\emph{Information Fusion}}  \bibinfo{volume}{45} (\bibinfo{year}{2019}), \bibinfo{pages}{227--245}.
\newblock


\bibitem[Seijo-Pardo et~al\mbox{.}(2017b)]%
        {seijo2017ensemble}
\bibfield{author}{\bibinfo{person}{Borja Seijo-Pardo}, \bibinfo{person}{Iago Porto-D{\'\i}az}, \bibinfo{person}{Ver{\'o}nica Bol{\'o}n-Canedo}, {and} \bibinfo{person}{Amparo Alonso-Betanzos}.} \bibinfo{year}{2017}\natexlab{b}.
\newblock \showarticletitle{Ensemble feature selection: homogeneous and heterogeneous approaches}.
\newblock \bibinfo{journal}{\emph{Knowledge-Based Systems}}  \bibinfo{volume}{118} (\bibinfo{year}{2017}), \bibinfo{pages}{124--139}.
\newblock


\bibitem[Sugumaran et~al\mbox{.}(2007)]%
        {sugumaran2007feature}
\bibfield{author}{\bibinfo{person}{V Sugumaran}, \bibinfo{person}{V Muralidharan}, {and} \bibinfo{person}{KI Ramachandran}.} \bibinfo{year}{2007}\natexlab{}.
\newblock \showarticletitle{Feature selection using decision tree and classification through proximal support vector machine for fault diagnostics of roller bearing}.
\newblock \bibinfo{journal}{\emph{Mechanical systems and signal processing}} \bibinfo{volume}{21}, \bibinfo{number}{2} (\bibinfo{year}{2007}), \bibinfo{pages}{930--942}.
\newblock


\bibitem[Thaseen and Kumar(2017)]%
        {chi-square}
\bibfield{author}{\bibinfo{person}{Ikram~Sumaiya Thaseen} {and} \bibinfo{person}{Cherukuri~Aswani Kumar}.} \bibinfo{year}{2017}\natexlab{}.
\newblock \showarticletitle{Intrusion detection model using fusion of chi-square feature selection and multi class SVM}.
\newblock \bibinfo{journal}{\emph{Journal of King Saud University-Computer and Information Sciences}} \bibinfo{volume}{29}, \bibinfo{number}{4} (\bibinfo{year}{2017}), \bibinfo{pages}{462--472}.
\newblock


\bibitem[Tibshirani(1996)]%
        {lasso}
\bibfield{author}{\bibinfo{person}{Robert Tibshirani}.} \bibinfo{year}{1996}\natexlab{}.
\newblock \showarticletitle{Regression shrinkage and selection via the lasso}.
\newblock \bibinfo{journal}{\emph{Journal of the Royal Statistical Society: Series B (Methodological)}} \bibinfo{volume}{58}, \bibinfo{number}{1} (\bibinfo{year}{1996}), \bibinfo{pages}{267--288}.
\newblock


\bibitem[Vaswani et~al\mbox{.}(2017)]%
        {vaswani2017attention}
\bibfield{author}{\bibinfo{person}{Ashish Vaswani}, \bibinfo{person}{Noam Shazeer}, \bibinfo{person}{Niki Parmar}, \bibinfo{person}{Jakob Uszkoreit}, \bibinfo{person}{Llion Jones}, \bibinfo{person}{Aidan~N Gomez}, \bibinfo{person}{{\L}ukasz Kaiser}, {and} \bibinfo{person}{Illia Polosukhin}.} \bibinfo{year}{2017}\natexlab{}.
\newblock \showarticletitle{Attention is all you need}.
\newblock \bibinfo{journal}{\emph{Advances in neural information processing systems}}  \bibinfo{volume}{30} (\bibinfo{year}{2017}).
\newblock


\bibitem[Yang and Honavar(1998)]%
        {yang1998feature}
\bibfield{author}{\bibinfo{person}{Jihoon Yang} {and} \bibinfo{person}{Vasant Honavar}.} \bibinfo{year}{1998}\natexlab{}.
\newblock \showarticletitle{Feature subset selection using a genetic algorithm}.
\newblock In \bibinfo{booktitle}{\emph{Feature extraction, construction and selection}}. \bibinfo{publisher}{Springer}, \bibinfo{pages}{117--136}.
\newblock


\bibitem[Yang and Pedersen(1997)]%
        {kbest}
\bibfield{author}{\bibinfo{person}{Yiming Yang} {and} \bibinfo{person}{Jan~O Pedersen}.} \bibinfo{year}{1997}\natexlab{}.
\newblock \showarticletitle{A comparative study on feature selection in text categorization}. In \bibinfo{booktitle}{\emph{Icml}}, Vol.~\bibinfo{volume}{97}. Nashville, TN, USA, \bibinfo{pages}{35}.
\newblock


\bibitem[Yu and Liu(2003)]%
        {yu2003feature}
\bibfield{author}{\bibinfo{person}{Lei Yu} {and} \bibinfo{person}{Huan Liu}.} \bibinfo{year}{2003}\natexlab{}.
\newblock \showarticletitle{Feature selection for high-dimensional data: A fast correlation-based filter solution}. In \bibinfo{booktitle}{\emph{Proceedings of the 20th international conference on machine learning (ICML-03)}}. \bibinfo{pages}{856--863}.
\newblock


\bibitem[Zhou and Wang(2007)]%
        {t-test}
\bibfield{author}{\bibinfo{person}{Nina Zhou} {and} \bibinfo{person}{Lipo Wang}.} \bibinfo{year}{2007}\natexlab{}.
\newblock \showarticletitle{A modified T-test feature selection method and its application on the HapMap genotype data}.
\newblock \bibinfo{journal}{\emph{Genomics, proteomics \& bioinformatics}} \bibinfo{volume}{5}, \bibinfo{number}{3-4} (\bibinfo{year}{2007}), \bibinfo{pages}{242--249}.
\newblock


\end{thebibliography}

\end{document}